\RequirePackage{fix-cm} 
\documentclass[final]{svjour3} 
\smartqed  
\makeatletter
\def\cl@chapter{\@elt {theorem}}
\makeatother

\usepackage{lipsum}
\usepackage{amsfonts}
\usepackage{graphicx}
\usepackage{epstopdf}
\usepackage{algorithmic}
\usepackage{float}
\usepackage{amsmath}
\usepackage{amssymb}
\usepackage{graphicx}
\usepackage{bm}
\usepackage[margin=1.2in]{geometry}

\usepackage{cleveref}

\usepackage{amsopn}

\DeclareMathOperator*{\argmin}{arg\,min}

\usepackage{xcolor}
\definecolor{color1th}{RGB}{140,182,154}
\definecolor{color2th}{RGB}{139,108,165}
\definecolor{color3th}{RGB}{112,227,246}
\definecolor{color4th}{RGB}{98,202,135}
\definecolor{color5th}{RGB}{145,236,18}
\definecolor{color6th}{RGB}{22,5,212}
\definecolor{color7th}{RGB}{198,222,250}
\definecolor{color8th}{RGB}{204,118,199}
\definecolor{color9th}{RGB}{30,163,37}
\definecolor{color10th}{RGB}{241,133,106}

\graphicspath{ {images/} }
\journalname{Noname}
\begin{document}

\title{Long-time predictive modeling of nonlinear dynamical systems using neural networks}
\subtitle{}


\author{Shaowu Pan          \and
        Karthik Duraisamy 
}


\institute{Shaowu Pan \at
              Department of Aerospace Engineering, University of Michigan, Ann arbor, MI  \\
              \email{shawnpan@umich.edu}           
           \and
           Karthik Duraisamy \at
             Department of Aerospace Engineering, University of Michigan, Ann arbor, MI  \\
              \email{kdur@umich.edu} 
}

\date{Received: date / Accepted: date}

\maketitle

\begin{abstract}
We study the use of feedforward neural networks (FNN) to develop models of non-linear dynamical systems from data. Emphasis is placed on predictions at long times, with limited data availability. Inspired by global stability analysis, and the observation of strong correlation between the local error and the maximal singular value of the Jacobian of the ANN, we introduce Jacobian regularization in the loss function. This regularization suppresses the sensitivity of the prediction to the local error and is shown to improve accuracy and robustness. Comparison between the proposed approach and sparse polynomial regression is presented in numerical examples ranging from simple ODE systems to nonlinear PDE systems including vortex shedding behind a cylinder, and instability-driven buoyant mixing flow. Furthermore, limitations of  feedforward neural networks are highlighted, especially when the training data does not include a low dimensional attractor. Strategies of data augmentation are presented as remedies to address these issues to a certain extent.
\keywords{data-driven modeling \and artificial neural networks \and dynamical system modeling \and machine learning}
\end{abstract}

\section{Introduction}

The need to model dynamical behavior from data is pervasive across science and engineering. Applications are found in diverse fields such as in control systems~\cite{tanaskovic2017data},  time series modeling~\cite{shumway2000time}, and describing the evolution of coherent structures~\cite{Brunton2015}. While data-driven modeling of dynamical systems can be broadly classified as a special case of system identification~\cite{mangan2016inferring}, it is important to note certain distinguishing qualities: the learning process may be performed off-line; physical systems may involve very high dimensions; and the goal may involve the prediction of long-time behaviour from limited training data. 

Artificial neural networks (ANN) have attracted considerable attention in recent years in domains such as image recognition in computer vision~\cite{russakovsky2015imagenet,karpathy2014large} and in control applications~\cite{duriez2017machine}. The success of ANNs arises from their ability to effectively learn low-dimensional representations from complex data, and in building relationships between features and outputs. Neural networks with a single hidden layer and nonlinear activation function are guaranteed to be able to predict any Borel measurable function to any degree of accuracy on a compact domain~\cite{hornik1991approximation}. 

The idea of leveraging neural networks to model dynamical systems has been explored  since the 1990s. ANNs are prevalent in the system identification and time series modeling community~\cite{Narendra1990,Polycarpou1991,Narendra1992,Kuschewski1993}, where the mapping between inputs and outputs is of prime interest. Billings et al.~\cite{Billings1992} explored connections between  neural networks and the nonlinear autoregressive moving average model (NARMAX) with exogenous inputs. It was shown that neural networks with one hidden layer and sigmoid activation function represents an infinite series consisting of polynomials of the input and state units. Elanayar et al.~\cite{Elanayar1994} proposed the approximation of nonlinear stochastic dynamical systems using radial basis feedforward neural networks. Early work using neural networks to forecast multivariate time series of commodity prices~\cite{Chakraborty1992} demonstrated its ability to model stochastic systems without knowledge of the underlying governing equations. Tsung et al.~\cite{tsung1995phase} proposed learning the dynamics in phase space using a feedforward neural network with time-delayed coordinates.

Paez and Urbina~\cite{paez1997dynamical,urbina1998characterization,paez2000nonlinear} modeled a nonlinear hardening oscillator using a neural network-based model combined with dimension reduction using canonical variate analysis (CVA).
Smaoui~\cite{Smaoui2001,Smaoui2004,Smaoui1997a} pioneered the use of neural networks to predict fluid dynamic systems such as the unstable manifold model for bursting behavior in the 2-D Navier-Stokes and the Kuramoto-Sivashinsky equations. The dimensionality of the original PDE system is reduced by considering a small number of Proper Orthogonal Decomposition (POD) coefficients~\cite{berkooz1993proper}. Interestingly, similar ideas of using principal component analysis for dimension reduction can be traced back to work in  cognitive science by Elman~\cite{elman1990finding}. Elman also  showed that knowledge of the intrinsic dimensions of the system can be very helpful in determining the structure of the neural network. However, in the majority of the results~\cite{Smaoui2001,Smaoui2004,Smaoui1997a}, the neural network model is only evaluated a few time steps from the training set,  which might not be a stringent performance test if longer time predictions are of interest. 

ANNs have also been applied to  chaotic nonlinear systems that are challenging from a data-driven modeling perspective, especially if long time predictions are desired. Instead of minimizing the pointwise prediction error, Bakker et al.~\cite{Bakker2000} satisfied the Diks' criterion in learning the chaotic attractor. Later, Lin et al.~\cite{Lin2003} demonstrated that, even the simplest feedfoward neural network for nonlinear chaotic hydrodynamics can show consistency in the time-averaged characteristics, power spectra, and Lyapunov exponent between the measurements and the model. 

A major difficulty in modeling dynamical systems is the issue of memory. It is known that even for a Markovian system, the corresponding reduced-dimensional system could be non-Markovian~\cite{chorinbook,Parish2016}. 
In general, there are  two main ways of introducing memory effects in neural networks. First, a simple workaround for feedforward neural networks (FNN) is to introduce time delayed states in the inputs~\cite{Narendra1992}.
However, the drawback is that this could potentially lead to an unnecessarily large number of parameters~\cite{Koskela1996}. To mitigate this, Bakker~\cite{Bakker2000} considered following Broomhead and King~\cite{broomhead1986extracting} in reducing the dimension of the delay vector using weighted principal component analysis (PCA). While the second approach  uses output or hidden units as additional feedback. As an example,  Elman's network~\cite{Koskela1996} is a recurrent neural network (RNN) that incorporates memory in a dynamic fashion.

Miyoshi et al. \cite{miyoshi1995learning} demonstrated that recurrent RBF networks have the ability to reconstruct simple chaotic dynamics. Sato et al. \cite{sato1996evolutionary} showed evolutionary algorithms can be used to train recurrent neural networks to capture the Lorenz system. Bailer-Jones et al. \cite{bailer1998recurrent} used a standard RNN to predict the time derivative in discrete or continuous form for simple dynamical systems; this can be considered an RNN extension to Tsung's phase space learning \cite{tsung1995phase}. 
Wang et al. \cite{Wang2017} proposed a framework combining POD for dimension reduction and long-short-term memory (LSTM) recurrent neural networks, and applied it to a fluid dynamic system. 


We limit ourselves to feedforward neural networks, since there are still many unanswered questions about modeling dynamical systems even in this simplest form. It is known that time delayed FNNs closely resemble  simple RNNs trained with teacher forcing~\cite{goodfellow2016deep}. 
Further, RNNs are not easy to train since  standard training algorithms (e.g., back propagation through time \cite{rumelhart1985learning}) are likely to introduce stronger overfitting than FNN due to  vanishing gradients \cite{goodfellow2016deep}. 
Recently, sparse regression (SINDy)~\cite{Brunton2015,mangan2016inferring} has gained popularity as a tool for data-driven modeling.  The idea is to  search for a sparse representation of a linear combination of functions selected from a library. In this work, we will compare it with FNN-based models and highlight some differences.

The paper is organized as follows: the problem description is provided in \cref{sec:prob_desc} and the mathematical formulation of standard and Jacobian-regularized FNNs is presented in \cref{sec:models}. Results and discussion are presented in \cref{sec:results}. We first present a comparison with SINDy for simple dynamical systems. Then we highlight the importance of stabilization to control the global error of predicted trajectory and the impact of Jacobian regularization. Finally, we apply the model in a nonlinear PDE system where a low dimensional attractor is not realized and discuss the limitations of black-box modeling of dynamical system and propose data augmentation as remedies. 
Conclusions are drawn in \cref{sec:conclusions}.



\section{Problem description}
\label{sec:prob_desc}

\textcolor{black}{
Consider a dynamical system in Euclidean space $\mathbb{R}^M$ which is described by a continuously differentiable function $\phi: \mathbb{R} \times \mathbb{R}^M \mapsto \mathbb{R}^M$, where $\phi(t, \bm{x}) = \phi_t(\bm{x})$. The state $\bm{x} \in \mathbb{R}^M$ satisfies the composition relation $\phi_{t+s} = \phi_t \circ \phi_s$ for $t,s \in \mathbb{R}$ and $\phi_0: \mathbb{R}^M \mapsto \mathbb{R}^M$ is the identity function. 
$\phi_t$ is the $t$ map of the flow described by a vector function $\bm{F}_{c}: \mathbb{R}^{M} \mapsto \mathbb{R}^M $ as 
\begin{equation}{\label{eq:transit_cont}}
 \dot{\bm x} =  \bm F_c( \bm x).
\end{equation}}

\textcolor{black}{
Similarly, one can define a discrete dynamical system induced by the above smooth dynamical system by considering a constant time step $\Delta t \in \mathbb{R}$ and a state transition map $\bm{F}_d(\bm{x}) = \phi_{\Delta t}(\bm{x}): \mathbb{R}^M \mapsto \mathbb{R}^M$ such that  
\begin{equation}{\label{eq:transit}}
\bm x^{n+1} =  \bm F_d( \bm x^{n}).
\end{equation}
Equivalently, one can rewrite the above system as
\begin{equation}{\label{eq:time_deri_2}}
(\bm{x}^{n+1} - \bm{x}^{n} )/\Delta t =  (\bm{F}_{d}( \bm x^{n}) - \bm{x}^n)/\Delta t \triangleq \bm{F}_r(\bm{x}^n),
\end{equation}
where $\bm{F}_r: \mathbb{R}^M \mapsto \mathbb{R}^M$ resembles a first order solution~\cite{bailer1998recurrent} to $\bm{F}_c$.
}

\textcolor{black}{
Our goal is to find an approximation to the dynamics, either in (i) a discrete sense $\bm{F}_d$, given data $\mathcal{D} = \{\bm{x}^i\}_{i=0}^{N-1}$ uniformly sampled from a trajectory given initial condition $\bm{x}^0 = \bm{x}(t=0)$, or (ii) in a continuous sense $\bm{F}_c$, given data $\mathcal{D} = \{ (\bm{x}^i, \dot{\bm{x}}^i ) \}_{i=0}^{N-1}$, where $N$ is the number of data points. It must be mentioned that - as highlighted in the result section - data does not have to be collected on the same trajectory.
}
%

Depending on the way one defines the training and testing set, two types of problems are considered in the current work: 
\begin{enumerate}
    \item  prediction of a certain trajectory starting from an initial condition that is different from the training trajectories.
    \item  prediction of the future trajectory given  past information of the trajectory as training data.
\end{enumerate}

Conservatively speaking, the success of tackling the first of the above problems requires the trajectories in the training data to be representative of the distribution in the region of interest, which may or may not be feasible depending on how informative the data is. In the context of modeling dynamical systems, it is often implied in previous literature~\cite{Smaoui2004} that the initial condition of unseen testing data is not far away from the training data. The second problem can also be difficult since it will challenge the effectiveness of the model  as past information might not be sufficient for the model to be predictive on unseen data. Again, it is often implied in previous works~\cite{Wang2017,trischler2016synthesis,Smaoui1997a}, successful predictions are often accompanied by an underlying low dimensional attractor so the past states as training data can be collected until it becomes representative of the future.

\section{Mathematical Modeling Framework}
\label{sec:models}

In this section, we first define performance metrics of the approximation to the dynamics \textcolor{black}{$f$}, then introduce the standard FNN model and the Jacobian-regularized FNN model. Finally,  techniques to mitigate overfitting  are described.

\subsection{Definitions of error metrics}

To measure the prediction error for each sample in an \emph{a priori} sense (i.e., given exact $\bm{x}^{i}$), we define the local error vector $\xi_{local}^{i} \in \mathbb{R}^M$ for the $i$-th sample $(\bm{x}^{i}, \bm{y}^{i})$ \textcolor{black}{as} 
\begin{equation}{\label{def:local_error_vector}}
\xi^i_{local} = \bm y^i - f(\bm x^{i}),
\end{equation}
\textcolor{black}{where $\bm{y}^i \in \mathbb{R}^{M}$ is the $i$-th target to learn from the $i$-th feature $\bm{x}^i \in \mathbb{R}^{M}$. For example, the feature is the state vector at $i$-th step, $\bm{x}^{i}$ and the target can be $\bm{x}^{i+1}$ for discrete dynamical system or $\bm{\dot{x}}^{i}$ for continuous dynamical system.}

Then, we can define local error at the $i$-th sample by 
\begin{equation}{\label{def:local_error}}
{e}_{local}^i = \lVert \xi^i_{local} \rVert_2 = \lVert \bm y^i - f(\bm x^{i}) \rVert_2,
\end{equation}
where $\lVert \cdot  \rVert_2: \mathbb{R}^M \mapsto [0, +\infty)$ is the vector 2-norm, i.e., $l^2$ norm, and $\lvert \cdot \rvert: \mathbb{R} \mapsto [0, \infty)$ is the absolute value. 

We can further define the local error of the $i$-th sample for the $j$-th component as shown by 
\begin{equation}{\label{def:local_error_j}}
{e}_{local,j}^i = \lvert y^i_j - f(\bm x^{i})_j \rvert.
\end{equation}

The local error assumes that the $i$-th input feature $\bm x_{i}$, is predicted accurately. 
On the other hand, the global error vector is defined by \cref{def:global_error_vector}, in which $\widehat{\bm x}^i$ is obtained by \emph{iterative prediction}, i.e., \emph{a posteriori evaluation}, at the $i$-th step from an initial condition through either time integration or transition function as a discrete map. That is, $\widehat{\bm x}^i$ is obtained from $f(\widehat{\bm x}^{i-1})$ in a recursive sense \textcolor{black}{as follows} 
\begin{equation}{\label{def:global_error_vector}}
\xi^i_{global} = \bm x^i - \widehat{\bm x}^i. 
\end{equation}

Similarly, the global error is defined by 
\begin{equation}{\label{def:global_error}}
{e}_{global}^i = \lVert \xi^i_{global} \rVert_2 = \lVert \bm x^i - \widehat{\bm x}^i \rVert_2,
\end{equation}
and for the $j$-th component specifically by 
\begin{equation}{\label{def:global_error_j}}
{e}_{global,j}^i = \lvert x^i_j - \widehat{x}^i_j \rvert.
\end{equation}

Further, to obtain a holistic \textcolor{black}{view of the} model performance in feature space, if $\bm{F}_d$ or $\bm{F}_c$ is known, \textcolor{black}{either in the continuous or discrete case}, we can define stepwise error as
\begin{equation}{\label{eq:step_error}}
    e_{stepwise}(\bm x) = \lVert \bm{F}_{c,d} (\bm x) - f (\bm x)  \rVert_2.
\end{equation}
Note that $e_{stepwise}$ is not restricted by the training or testing trajectory, but \textcolor{black}{it can be evaluated arbitrarily} in the region of interest.

\textcolor{black}{
Finally,  we consider the uniform averaged coefficient of determination $R^2$ as a scalar metric for measuring regression performance 
\begin{equation}{\label{eq:R2}}
R^2 = \frac{1}{M} \sum_{j=0}^{M-1} R^2_j,
\end{equation}
}
\textcolor{black}{where $R^2_j$ is given by}
\begin{equation}
    R^2_j = R^2(\bm y_j, \widetilde{\bm y}_j ) = 1 - \frac{\sum_{i=0}^{n_{sample}-1} (y^i_j - \widetilde{y}^i_j )^2 }{\sum_{i=0}^{n_{sample}-1} (y^i_j - \bar{y}_j)^2},
\end{equation}
where \textcolor{black}{$n_{sample}$ is the number of samples in the validation data}, $\bar{ y}_j = \frac{1}{n_{sample}} \sum_{i=0}^{n_{sample}-1}  y^i_j$ and 
$\widetilde{\bm y}^i = f(\bm x^{i})$ is the prediction of $f$ based on $i$-th feature $\bm x^{i}$.


\subsection{Feedforward neural network model}

\subsubsection{Basic model: densely connected feedforward neural network}

\textcolor{black}{The basic model approximates $\bm{F}_c$ in \cref{eq:transit_cont} for the continuous case, and $\bm{F}_r$ in \cref{eq:time_deri_2} in the discrete case using a feedforward neural network. } The existence of an arbitrarily accurate feedforward neural network approximation to any Borel measureable function given enough number of hidden units is guaranteed from the property of the universal approximator~\cite{hornik1989multilayer}. It should be noted that our basic model is related to Tsung's phase-space-learning model \cite{tsung1995phase}. If the Markovian assumption is adopted, the training feature matrix snapshots $X$ and training target matrix snapshots $Y$ are as follows:
\begin{equation}
X = 
\begin{bmatrix}
x^0_1       & x^0_2 & x^0_3 & \dots & x^0_{M} \\
x^1_1       & x^1_2 & x^1_3 & \dots & x^1_{M} \\
\hdotsfor{5} \\
x^{N-1}_{1}      & x^{N-1}_2 & x^{N-1}_3 & \dots & x^{N-1}_{M} \\
\end{bmatrix}
\in \mathbb{R}^{N \times M},
\end{equation}
and 
\begin{equation}
Y = 
\begin{bmatrix}
y^0_1       & y^0_2 & y^0_3 & \dots & y^0_{M} \\
y^1_1      & y^1_2 & y^1_3 & \dots & y^1_{M} \\
\hdotsfor{5} \\
y^{N-1}_{1}      & y^{N-1}_2 & y^{N-1}_3 & \dots & y^{N-1}_{M} \\
\end{bmatrix}
\in \mathbb{R}^{N \times M},
\end{equation}
where $M$ is the dimension of the state, $N$ is the total number of snapshots of training data, learning target $Y$ is the time derivative, and the subscript stands for the index of the component. Note that each component of the feature and target are normalized to zero mean and unit variance for better training performance in the neural network.

By generally constructing a densely connected feedforward neural network $f(\cdot)$: $\mathbb{R}^M \mapsto \mathbb{R}^M$ with $L-1$ hidden layers and output layer as linear, the following recursive expression is defined for each hidden layer:
\begin{equation}{\label{eq:recursive}}
\bm \eta ^l = \sigma_l (W_l \bm \eta^{l-1} + b_l), \ \ l=1,\dots,L-1,
\end{equation}
 where $\eta^{0}$ stands for the input of the neural network $\bm x$, $\bm \eta^l \in \mathbb{R}^{n_l}$, $n_l \in \mathbb{N^{+}}$ is the number of hidden units in layer $l$, and $\sigma_l$ is the activation function of layer $l$.

Note that the output layer is linear, i.e., $\sigma_L(x)=x$:
\begin{equation}
 f(\bm x; \bm W^L, \bm b^L) = \bm \eta^L = W_L \bm \eta^{L-1} + b_L ,
 \end{equation}
where parameters of the neural network are $\bm W^L = \{ W_i \}_{i=1,\dots,L}$, $\bm b^L = \{ b_i \}_{i=1,\dots,L}$.

For example, if we consider using two hidden layers where $L=3$ and the number of hidden units are the same, the full expression for the neural network model is given by
\begin{equation}{\label{eq:full_basic_model}}
\hat{\bm{y}} = f(\bm x;\bm W, \bm b) = W_3\sigma(W_2 \sigma(W_1 \bm x + b_1 ) + b_2) + b_3,
\end{equation}
where $\bm x \in \mathbb{R}^M$ is the state of the dynamical system, \textcolor{black}{i.e., the input to the neural network} and $\hat{\bm{y}}\in \mathbb{R}^M$ is the modeling target, \textcolor{black}{ i.e., the output of neural network}. $\sigma(\cdot)$: $\mathbb{R} \mapsto \mathbb{R}$ is a nonlinear activation function. $W_1 \in \mathbb{R}^{ n_h \times M }$, $W_2 \in \mathbb{R}^{ n_h \times n_h }$, $W_3 \in \mathbb{R}^{ M \times n_h }$. $b_1 \in \mathbb{R}^{n_h}$, $b_2 \in \mathbb{R}^{n_h}$, $b_1 \in \mathbb{R}^{M}$. Sets of weights and biases are $\bm W^3 = \{W_1,W_2,W_3\}$ and $\bm b^3 = \{b_1,b_2,b_3\}$. \textcolor{black}{The problem is to find the set of parameters of $\bm{W}^3$ and $\bm{b}^3$ that result in the best approximation of the underlying ground truth ($\bm{F}_c, \bm{F}_d  \textrm{ or } \bm{F}_r$). Under the framework of statistical learning, it is standard to perform empirical risk minimization (ERM) with mean-square-error loss.
The set up and parameters corresponding to the desired solution $f(\bm{x}, \bm{W}^*, \bm{b}^*)$ can be written as} 
\begin{equation}{\label{eq:basic_model}}
\displaystyle \bm{W}^{*}, \bm{b}^{*} = \argmin_{\bm W^3, \bm b^3} \frac{1}{|I_{train}|} \sum_{i\in I_{train}} \lVert f(\bm x^i; \bm W^3, \bm b^3) - \bm y^i \rVert_2^2,
\end{equation}
where $I_{train}$ is the index set of training data, and $\bm x^i$ and $\bm y^i$ correspond to the $i$-$th$ feature-target pair. 

\textcolor{black}{To deal with the high dimensionality of the optimization problem, we employ Adam~\cite{kingma2014adam}, a gradient-based algorithm, which is essentially a mixture between momentum acceleration and rescaling parameters.
 The weights are initialized using a truncated normal distribution to potentially avoid saturation and use the Automatic Differentiation (AD) provided by Tensorflow~\cite{tensorflow} to compute the gradients. The neural network model is implemented in Python using the Tensorflow library~\cite{tensorflow}.}
Due to the non-convex nature of \cref{eq:basic_model}, for such a high degree of freedom of parameters, one can only afford to find a local minimum. In practice, however, a good local minimum is usually satisfactory \cite{goodfellow2016deep}. 
Hyperparameters considered in current work for the basic model are the number of units for each hidden layer $n_h$ and activation function $\sigma(\cdot)$. 

\textcolor{black}{Model selection for neural networks is an active research area~\cite{bergstra2012random,snoek2012practical,bergstra2013making}. 
Well known methods involve grid search/random search~\cite{bergstra2012random}/Tree of Parzen Estimators (TPE)~\cite{bergstra2013making}/Bayesian optimization~\cite{snoek2012practical} with cross validation. 
We pursue the following trial-error strategy:
\begin{enumerate}
\item Given the number of training points, computing the number of equations to satisfy if the network overfits all the training data.
\item Pick a neural network with uniform hidden layer structure to overfit the training data with the number of parameters in the network no more than 10\% to 50\% of the number in step 1.
\item Keep reducing the size of neural network by either decreasing the hidden units or number of layers until the training and validation error are roughly the same order.
\item For the choice of other hyperparameters, we simply perform grid search.
\end{enumerate}
}

\subsubsection{Jacobian regularized model}

In standard FNNs, minimizing mean-squared-error on the training data 
only guarantees model performance in terms of the local training error. It does not guarantee \textcolor{black}{the reconstruction of even training trajectory in the \emph{a posteriori sense}.}

Here, we take a closer look at the error propagation in a dynamical system for the FNN model when evaluated in an iterative fashion, \textcolor{black}{i.e., \emph{a posteriori sense}}. Without any loss of generality, \textcolor{black}{considering the discrete case, after we obtain the model $f$, we can predict $\widehat{\bm{x}}^{i+1}$ given $\widehat{\bm{x}}^i$}
\begin{equation}
   \widehat{\bm x}^{i+1} = f ( \widehat{\bm x}^{i} ),
\end{equation}

Moreover, \textcolor{black}{given $\bm{F}_d$, we can find the $\xi^{i+1}_{global}$ given $\widehat{\bm{x}}^i$ and $\xi^{i}_{global}$ as follows, } 
\begin{equation}
   \xi^{i+1}_{global} = {\bm x}^{i+1} - \widehat{\bm x}^{i+1}  = \bm{F}_d ( {\bm x}^{i} ) - f( \widehat{\bm x}^{i} ) = \bm{F}_d ( \widehat{\bm x}^{i} + \xi^{i}_{global} ) - f( \widehat{\bm x}^{i} ),
\end{equation}
\begin{equation}
   \xi^{i+1}_{global} = \bm{F}_d( \widehat{\bm x}^{i} + \xi^{i}_{global} ) -f( \widehat{\bm x}^{i} + \xi^{i}_{global} ) + f( \widehat{\bm x}^{i} + \xi^{i}_{global} ) - f( \widehat{\bm x}^{i} ).
\end{equation}

Consider a Taylor expansion of $f( \widehat{\bm x}^{i} + \xi^{i}_{global} )$ about $ \widehat{\bm x}^{i}$, we have
\begin{equation}
    \xi^{i+1}_{global} = \bm{F}_d( \widehat{\bm x}^{i} + \xi^{i}_{global} ) -f( \widehat{\bm x}^{i} + \xi^{i}_{global} ) 
    + \left.\dfrac{\partial f}{\partial \bm x}\right\vert_{\bm x = \widehat{\bm x}^{i} } \xi^{i}_{global} + \frac{1}{2}{\xi^{i}_{global}}^{T} H \xi^{i}_{global} + \ldots,
\end{equation}
where $H$ is the Hessian matrix evaluated at some point between $\widehat{\bm x}^{i}$ and $\widehat{\bm x}^{i} + \xi^{i}_{global}$. 

Assuming $\lVert \xi^{i}_{global}\rVert_2 \ll 1$, $\lVert H \rVert_2$ is bounded, and the high order terms are negligible compared to the Jacobian term we have  
\begin{equation}{\label{eq:ex_ineq}}
e^{i+1}_{global} 
\le e^{i}_{local} + \left \lVert \left. \dfrac{\partial f}{\partial \bm x}\right\vert_{\bm x = \widehat{\bm x}^{i} } \right \rVert_2 e^{i}_{global} + o(e^{i}_{global}) 
\le 
e^{i}_{local} + \left \lVert \left.\dfrac{\partial f}{\partial \bm x}\right\vert_{\bm x = \widehat{\bm x}^{i} } \right \rVert_F e^{i}_{global} + o(e^{i}_{global}).
\end{equation}

Similarly, \textcolor{black}{in the continuous case}, we have 
\begin{equation}
    e^{i+1}_{global} \le e^{i}_{global} + \int_{i\Delta t}^{(i+1)\Delta t} e^{\tau/\Delta  t}_{local} d\tau + \int_{i\Delta t}^{(i+1)\Delta t} \left \lVert \left . \dfrac{\partial f}{\partial \bm x}\right\rvert_{\bm x = \widehat{\bm x}^{i} } \right \rVert_F e^{\tau/\Delta t}_{global} d \tau + o(e^{i}_{global}\Delta t),
\end{equation}
\begin{equation}{\label{eq:td_ineq}}
e^{i+1}_{global} \le  \left( 1 + \Delta t \left \lVert \left . \dfrac{\partial f}{\partial \bm x}\right\rvert_{\bm x = \widehat{\bm x}^{i} } \right \rVert_F \right) e^{i}_{global}  + \int_{i\Delta t}^{(i+1)\Delta t} e^{\tau/\Delta  t}_{local} d\tau + o(e^{i}_{global}\Delta t).
\end{equation}

The right hand sides of \cref{eq:ex_ineq} and \cref{eq:td_ineq} contain contributions from the global error and accumulation of local error. Optimization as in \cref{eq:basic_model} can minimize the latter term, but not necessarily the former. This suggests that manipulating the eigenspectrum of the Jacobian might be beneficial for stabilization by suppressing the growth of the error. Due to the simplicity of computing the Frobenius norm compared to the 2-norm, we consider penalizing the Frobenius norm of the Jacobian of the neural network model. In the context of improving generalization performance of input-output neural network models, similar regularization has been also proposed by Rifai~\cite{rifai2011adding}. It should be noted that our purpose is to achieve better error dynamics in a temporal sense, which differs from the generalization goal in  deep learning. Thus, one may seek a locally optimal solution that can suppresses the growth in global error while minimizing the local error. 

The regularized loss function inspired from the above discussion is thus 
\begin{equation}{\label{eq:regu_model}}
\displaystyle \bm{W}^{*},\bm{b}^{*} = \argmin_{ \bm W,\bm b} \frac{1}{N_{train}} \sum_{i\in I_{train}} \lVert f(\bm x^i; \bm W,\bm b) - \bm y^i \rVert_2^2  +  \lambda
\lVert J(\bm x^i; \bm W,\bm b)\rVert_F^2,
\end{equation}
where $J$ is the Jacobian of the neural network output with respect to the input, and $\lambda$ is a hyperparameter. On one hand, it should be noted that regularizing the Frobenius norm of the eigenspectrum of the Jacobian  indirectly suppresses the magnitude of the eigenvalue of the Jacobian. On the other hand, excessive weighting on the magnitude of the eigenvalue would lead to less weighting on local error, which might result in an undesirably  large local error. Thus, $\lambda$ should be set as a relatively small value without strongly impacting the model performance in an a priori sense. 

\subsection{Reducing overfitting}
Overfitting is a common issue in the training of machine learning models, and it arises when models tend to memorize the training data instead of generalizing true functional relations. In neural networks, overfitting can occur from poor local minima and is partially due to the unavoidable non-convexity of an artificial neural network. Overfitting cannot be completely eliminated for most problems, given the NP-hard  nature of the problem. Generally, overfitting can be controlled by three kinds of regularization techniques: The first follows the Occam's razor principle, e.g., L1 sparsity regularization \cite{Brunton2015}. However there is no guarantee that Occam's razor is appropriate for all cases, and  finding the optimal sparsity level is often iterative. The second  is to smooth the function, e.g., using weight decay \cite{goodfellow2016deep}. The third type is especially suitable in iterative learning, e.g., early stopping, which is a widely used strategy in the deep learning community \cite{goodfellow2016deep}. In this work, we found validation-based early stopping to be sufficient. We split the  data further into pure training and validation sets, and then monitor overfitting by measuring $R^2$. 

\section{Results \& Discussion}
\label{sec:results}

Given sequential training data, the capability of the basic FNN is first evaluated in two-dimensional dynamical systems with polynomial non-linearities in \cref{sec:result_vdp} and non-polynomial-non-rational dynamics in \cref{sec:yg}. The basic model is compared with SINDy~\cite{Brunton2015}, a method that directly aims to learn functional models using $L_1$ sparse regression on a dictionary of candidate basis functions. In \cref{sec:cyd}, we demonstrate that the basic model performs better than SINDy on the problem of incompressible flow behind a cylinder, in spite of  the explicit addition of quadratic terms to the dictionary. In addition,  the local error is found  to be strongly correlated with the maximal singular value of the Jacobian, thus serving as an inspiration for Jacobian regularization. In \cref{sec:jacobian}, we demonstrate the stabilizing aspect of Jacobian regularization for the problem of laminar wake behind a cylinder, where the system exhibits a low dimensional attractor. In \cref{sec:buoymix}, we assess the ability of our regularized FNN model to approximate a dynamically evolving high-dimensional buoyancy-driven mixing flow system that is characteristic of flow physics driven by instabilities. The results show that, for systems that do not exhibit a low dimensional attractor, it is difficult for a black-box model to have satisfactory long-time prediction capabilities. In \cref{sec:info_rich} we show that predictive properties can be improved by data augmentation in the state space of interest. 



\subsection{2D polynomial system: Van der Pol oscillator}\label{sec:result_vdp}

The first order forward discretized scheme of the Van der Pol (VDP) system is given by 
\begin{equation}{\label{eq:vbe_first_order}}
\begin{pmatrix} x_1^{n+1}  \\ x_2^{n+1} \end{pmatrix} =  \begin{pmatrix}x_1^{n} 
\\ x_2^{n}
\end{pmatrix}+ \Delta t \begin{pmatrix}   x_2^{n} \\ (\mu (1-x_1^{n}x_1^{n})x_2^{n} - x_1^{n}) \end{pmatrix},
\end{equation} where $\Delta t = 0.1$ and $\mu = 2.0$. 
The modeling target is
\begin{equation}{\label{eq:vbe_first_order2}}
\bm y^n =
\begin{pmatrix} (x_1^{n+1} - x_1^{n})/\Delta t  \\ (x_2^{n+1} - x_2^{n})/\Delta t \end{pmatrix} = \begin{pmatrix}   x_2^{n} \\ (\mu (1-x_1^{n}x_1^{n})x_2^{n} - x_1^{n}) \end{pmatrix} 
 = \bm F_r(\bm x^n).
\end{equation}

\begin{figure}[!htb]
    \centering
    \includegraphics[width=\textwidth]{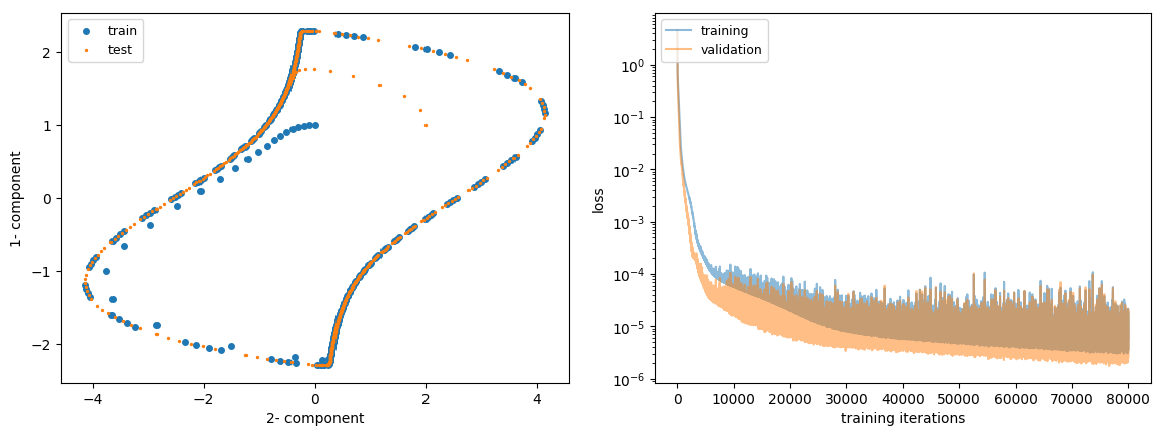}
	\caption{VDP case. Left: Distribution of training/testing sets. Right: learning rate.}
	\label{fig:vdp_train_target_dist_and_lc}
\end{figure}
Our goal is to reproduce the dynamics governed by $\bm F_r$ from on data collected from a single trajectory. 
The loss function of basic model in \cref{eq:basic_model} is optimized using training data from a single trajectory, containing 399 data points. Test data containing 599 points is generated using a different initial condition. The data distribution of the training  and testing features is shown in \cref{fig:vdp_train_target_dist_and_lc}. Note that initially a few test points (orange) are away from the training data (blue) which require the model to perform extrapolation.  
Configuration of hyperparameters is shown in \cref{table:hyperParameter_vdp_ann}. Data is normalized to zero mean and unit standard deviation for each component. We use mini-batch training with batch size = 64 and 80000 epochs.
\begin{table}[!htb]
    \caption{Hyperparameter configuration of the basic model: VDP case}
    \label{table:hyperParameter_vdp_ann}
    \centering
    \begin{tabular}{|c|c|c|c|c|} \hline
            layer structure & activation function & loss function & optimizer & learning rate  \\ \hline
    		2-8-8-2 & Swish & MSE & Adam & 0.002 \\ \hline
    \end{tabular}
\end{table}
The basic model consists of two hidden layers with each layer containing 8 hidden units. Two hidden layers are accompanied by Swish nonlinear activation as $\sigma(x) = x \cdot{} \textrm{sigmoid}(\beta x)$ where in practice $\beta$ is fixed as unity \cite{swish2018}. The output layer is linear. Randomly 20\% of training data is used as a validation set and we monitor the performance on the validation set as a warning of overfitting. 
In \cref{fig:vdp_train_target_dist_and_lc}, \textcolor{black}{the learning curve suggests that  the model is well-trained and overfitting is not observed.}

\begin{figure}[!htb]
	\centering
	\includegraphics[width=\textwidth]{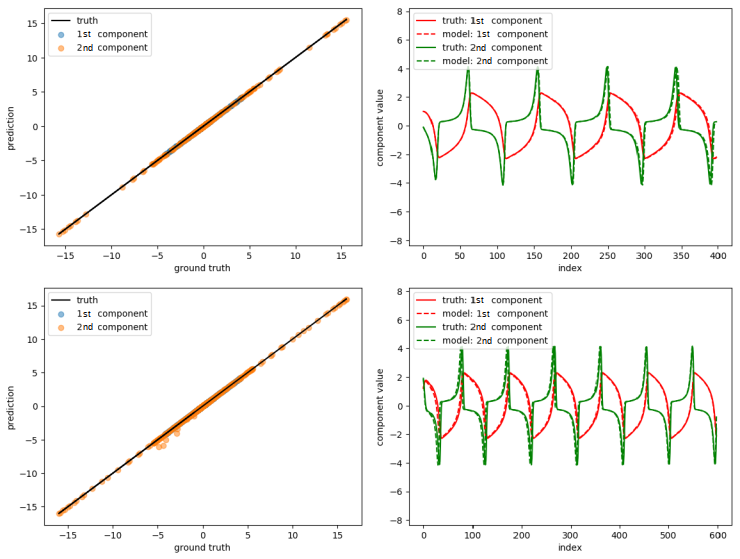}
	\caption{A priori and a posteriori result of basic model on VDP case. Top: training data. Bottom: testing data. Left: a priori. Right: a posteriori.}
	\label{fig:vdp_validation_0}
\end{figure}

Results of a priori and a posteriori prediction are shown in \cref{fig:vdp_validation_0}. The basic model predicts the $\bm F_r$ at each training point very well a priori, but slight phase lag is observed a posteriori in testing, which originates from the extrapolation of the testing data initially.

\begin{figure}[!htb]
	\centering
	\includegraphics[width=\textwidth]{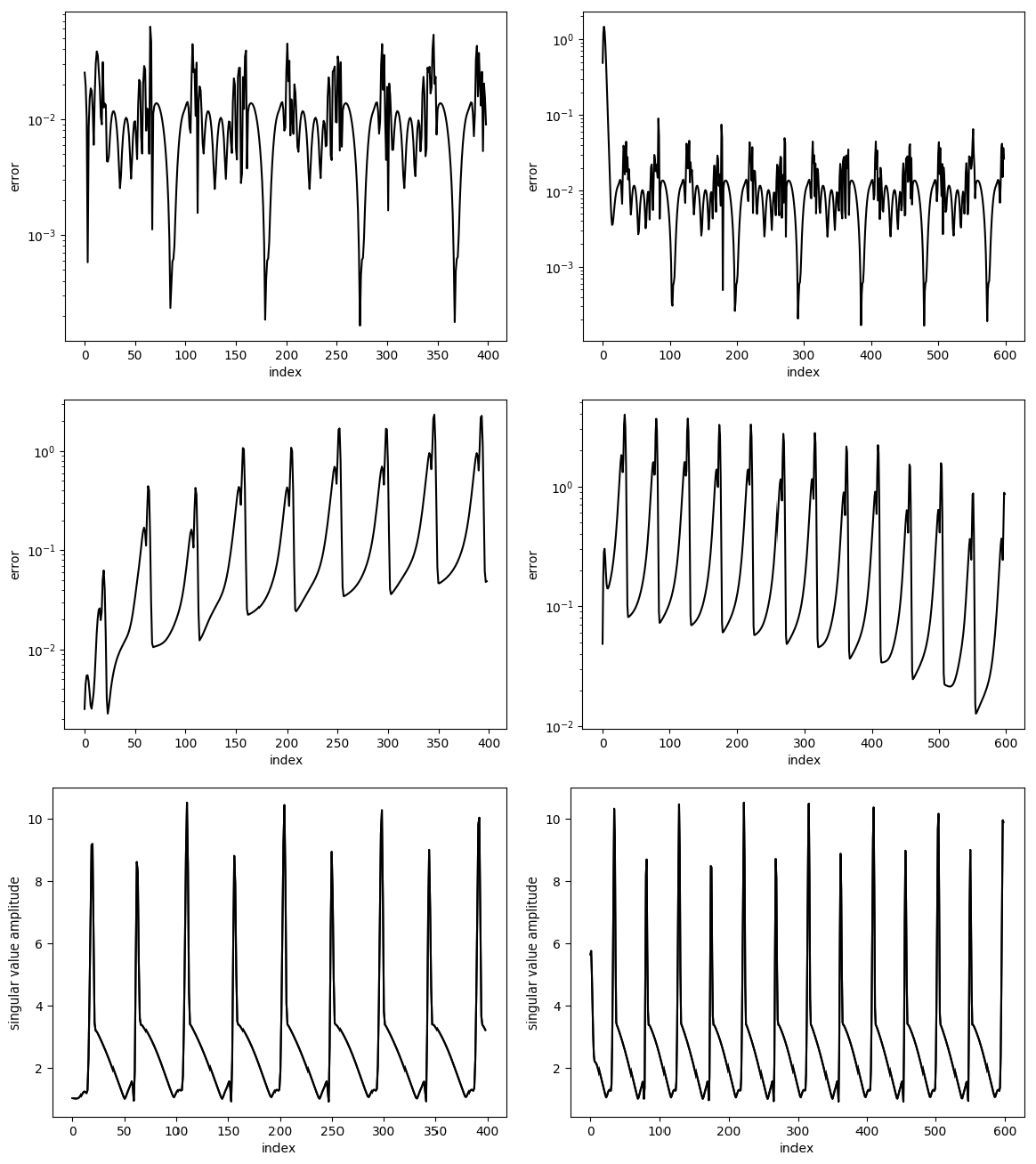}
	\caption{Variation of local error and global error: basic model on VDP case. Top: local error. Middle: global error. Bottom: maximal singular value of Jacobian evaluated in a priori. Left: training data. Right: testing data.}
	\label{fig:vdp_error}
\end{figure}

The variation of the local and global error  together with the maximal singular value of the Jacobian is shown in  \cref{fig:vdp_error}. For training data,  $e_{local}$ is observed to be relatively uniform, as expected since the objective optimized is MSE uniformly across all training data points. For testing data, $e_{local}$ exhibits peak values near the beginning of the trajectory as expected, since the first few points are far away from the training data shown in \cref{fig:vdp_train_target_dist_and_lc}. Moreover, it is interesting to observe that in \cref{fig:vdp_error}, the peak of the temporal history of local/global error shows strong correlation with the maximal singular value of the Jacobian.  

\begin{figure}[!htb]
	\centering
	\includegraphics[width=\textwidth]{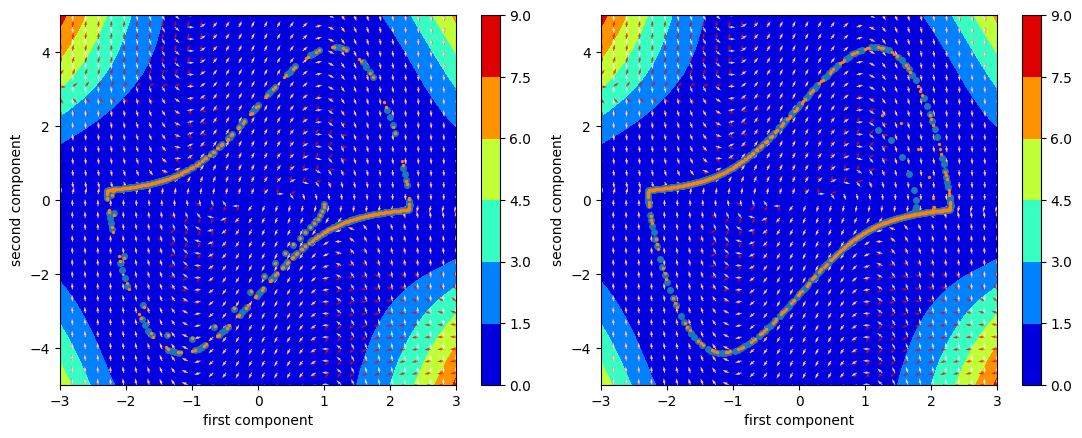}
	\caption{Stepwise error contour of basic model on VDP case. Left: training data. Right: testing data. Blue dot: ground truth. Orange dot: prediction. White arrow: direction of target vector of ground truth. Red arrow: direction of target vector of prediction.}
	\label{fig:error_contour_Vdp}
\end{figure}

Stepwise error contours are displayed in \cref{fig:error_contour_Vdp}. The region of large error close to red (implying the difference of the step-wise vector between neural network prediction and ground truth is large) is located near the corner of figure, where there is a dearth of training points. The model performs well near the training points as expected. In this case, since testing data is not very far away from the training data, good performance of extrapolation can be expected. \textcolor{black}{However, we would like to note that there is a moderate amount of error associated with the vector direction in \cref{fig:error_contour_Vdp} not only at the corners but also near the origin. This implies that a feedforward neural network can generalize to some extent, but with no guarantees, even in regions enclosed by training data. The results also confirm that the known result that for a dynamical system with an attractor, the neural network can reproduce the dynamics near the attractor~\cite{bailer1998recurrent,yulearning,Billings1992,Bakker2000,trischler2016synthesis}}.

\begin{figure}[!htb]
	\centering
	\includegraphics[width=\textwidth]{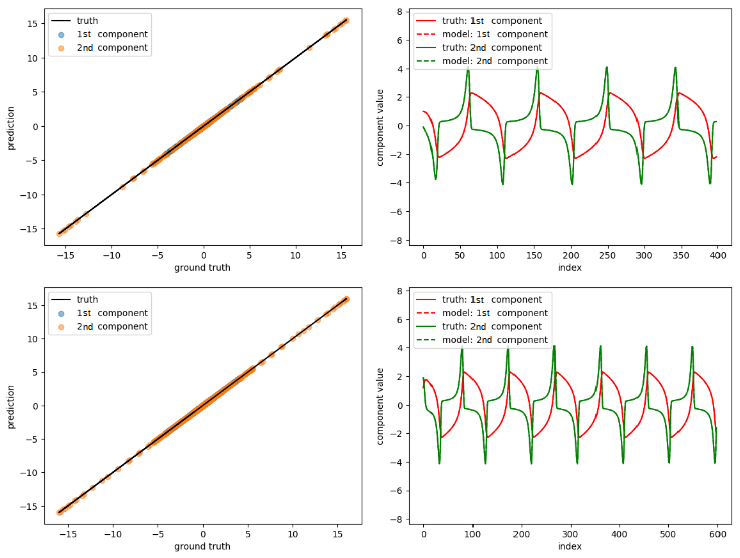}
	\caption{A priori and a posteriori result of SINDy on VDP case. Top: training data. Bottom: testing data. Left: a priori. Right: a posteriori.}
	\label{fig:vdp_validation_0_sindy}
\end{figure}

\begin{figure}[!htb]
	\centering
	\includegraphics[width=\textwidth]{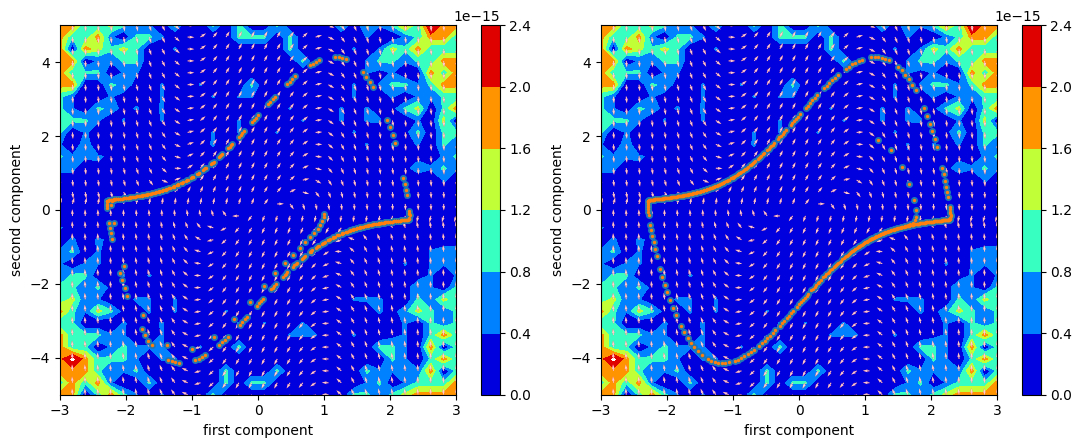}
	\caption{Stepwise error contour of SINDy on VDP case. Left: training data. Right: testing data. Blue dot: ground truth. Orange dot: prediction. White arrow: direction of target vector of ground truth. Red arrow: direction of target vector of prediction.}
	\label{fig:error_contour_Vdp_sindy}
\end{figure}

\textcolor{black}{With the prior knowledge that the system is polynomial in nature, one can use polynomial basis functions to extract the ground truth.} To illustrate this, results obtained from SINDy \cite{Brunton2015} are shown in \cref{fig:vdp_validation_0_sindy}, with threshold parameter as $2\times10^{-4}$, maximal polynomial order as 3, and no validation data set considered. As displayed in \cref{fig:error_contour_Vdp_sindy}, the excellent result of SINDy shows the advantage of finding the global features where parameters obtained are not restricted to the scope of training data since the ground truth is governed by sparse polynomials. 

\subsection{2D non-polynomial system: a non-rational non-polynomial oscillator}{\label{sec:yg}}

The success of SINDy is a consequence of the fact that the underlying system can be represented as a sparse vector in a predefined basis library such as that consisting of polynomial or rational functions \cite{mangan2016inferring}. Here, we choose a different case: a non-rational, non-polynomial oscillator with $\Delta t = 0.004$:

\begin{equation}{\label{eq:yg}}
\begin{pmatrix} x_1^{n+1}  \\ x_2^{n+1} \end{pmatrix} =  \begin{pmatrix}x_1^{n} 
	\\ x_2^{n}
	\end{pmatrix}+ \Delta t \begin{pmatrix}   2.5 - 100 \dfrac{x_1^{n}x_2^{n}}{1 + (x_2^{n}/0.52)^4} \\ -200 \dfrac{x_1^{n}x_2^{n}}{1 + (x_2^{n}/0.52)^4} + 9.2 - 2.3 x_2^{n} - 1.28|x_2^{n}|^{3/2} \end{pmatrix}
\end{equation}

\begin{figure}[!htb]
	\centering
	\includegraphics[width=\textwidth]{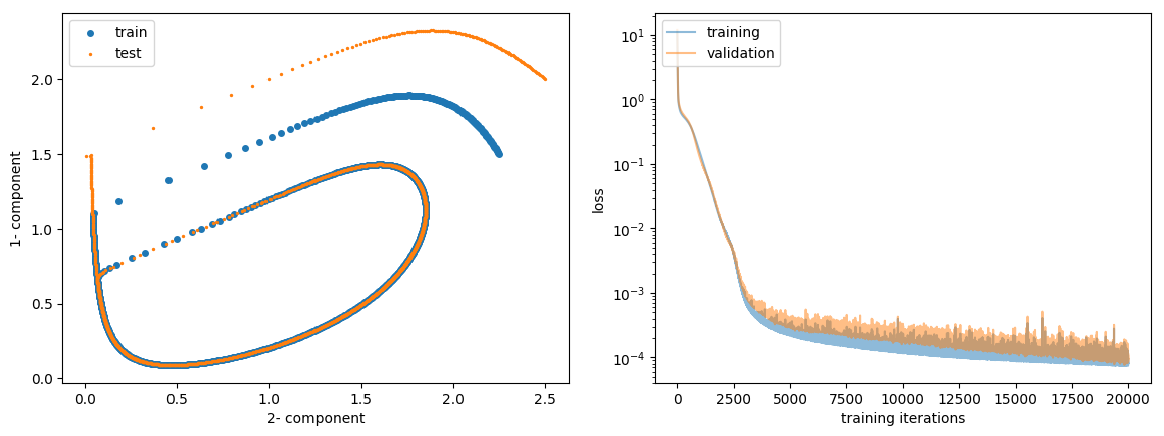}
	\caption{Non-rational non-polynomial case. Left: data distribution of training/testing feature. Right: learning curve.}
	\label{fig:yg_train_target_dist_and_lc}
\end{figure}

Here the basic model in \cref{eq:basic_model} is optimized using 1199 data points of a single trajectory. Testing data contains 1799 points. Randomly 20\% of training data is taken as the validation set, but also included in later evaluation. The feature distribution in phase space  is shown in \cref{fig:yg_train_target_dist_and_lc}. 
Hyperparameters are listed in  \cref{table:hyperParameter_yg_ann} and  128 minibatchs and 20000 epochs are used. The training error and validation error is also shown in \cref{fig:yg_train_target_dist_and_lc}. 

\begin{table}[!htb]
\caption{Hyperparameter configuration of basic model: non-rational non-polynomial case}
\label{table:hyperParameter_yg_ann}
\centering
\begin{tabular}{|c|c|c|c|c|} \hline
		layer structure & activation function & loss function & optimizer & learning rate  \\ \hline
		2-8-8-2 & elu & MSE & Adam & 0.005 \\  \hline
\end{tabular}
\end{table}

\begin{figure}[!htb]
	\centering
	\includegraphics[width=\textwidth]{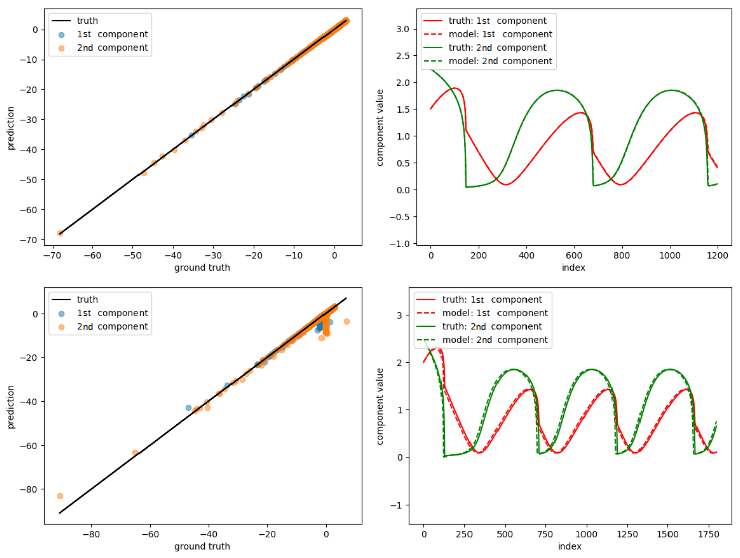}
	\caption{A priori and a posteriori result of basic model on non-rational non-polynomial case. Top: training data. Bottom: testing data. Left: a priori. Right: a posteriori.}
	\label{fig:yg_validation_0}
\end{figure}

Results for a priori and a posteriori performance on training and testing data are shown below in \cref{fig:yg_validation_0}. The training trajectory is perfectly reconstructed while the predictions show slight deviation.

\begin{figure}[!htb]
	\centering		\includegraphics[width=\textwidth]{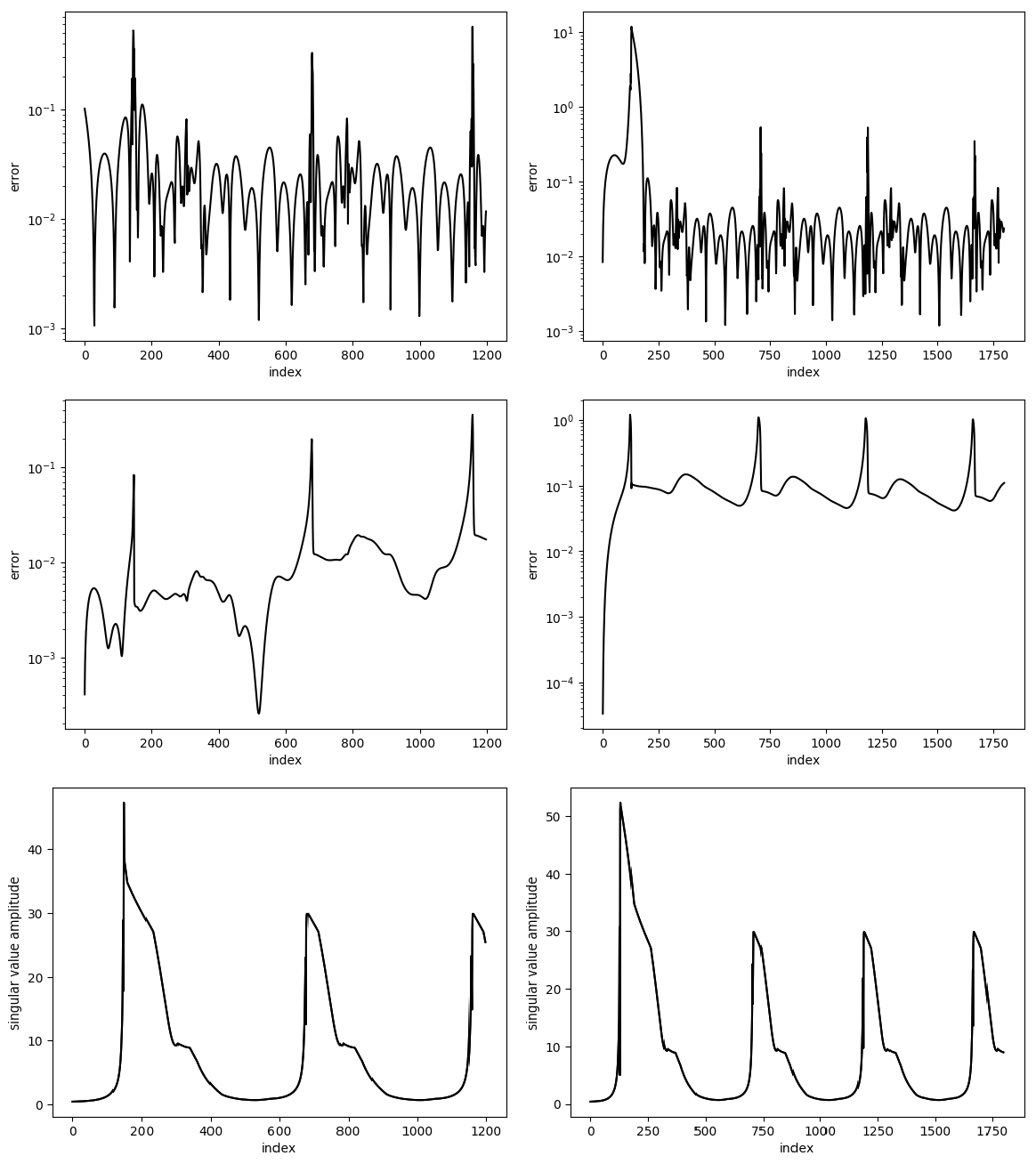}
    \caption{Variation of local error and global error: basic model on non-rational, non-polynomial case. Top: local error. Middle: global error. Bottom: maximal singular value of Jacobian evaluated in a priori. Left: training data. Right: testing data.}
	\label{fig:yg_error}
\end{figure}

The distribution of the local and global error is shown in \cref{fig:yg_error}. Again, we observe that maximal local/global error correlates with the peaks of the maximal singular value of the Jacobian. It is interesting to note that the highest local testing error occurs at the peak of the maximal singular value of the Jacobian, instead of at points close to the initial condition.

\begin{figure}[!htb]
	\centering
	\includegraphics[width=\textwidth]{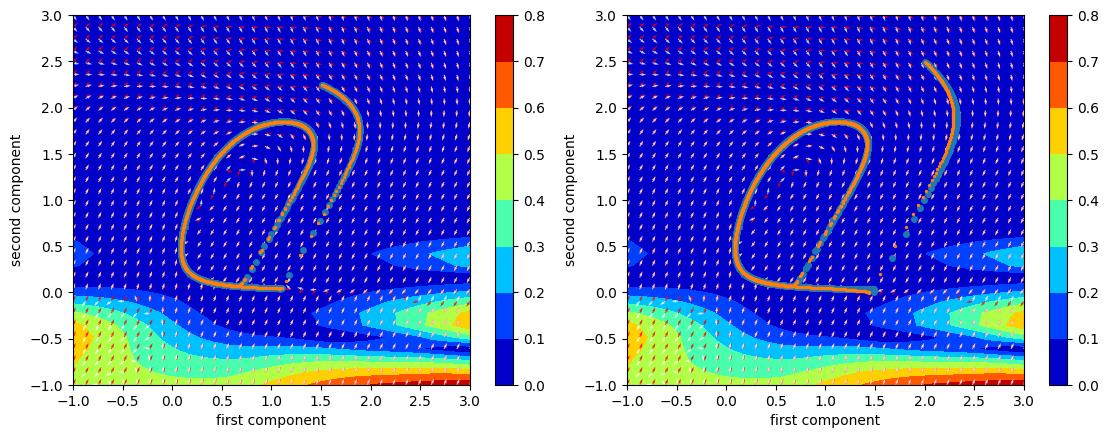}
	\caption{Stepwise error contour of basic model on non-rational, non-polynomial case. Left: training data. Right: testing data. Blue dot: ground truth. Orange dot: prediction. White arrow: direction of target vector of ground truth. Red arrow: direction of target vector of prediction.}
	\label{fig:error_contour_YG}
\end{figure}

The error contour in \cref{fig:error_contour_YG} shows that stepwise error around training trajectory is below 0.1. It is important to note that model performance deteriorates at places far away from the training trajectory, especially at the right corner shown in \cref{fig:error_contour_YG}.  

\begin{figure}[!htb]
	\centering
	\includegraphics[width=\textwidth]{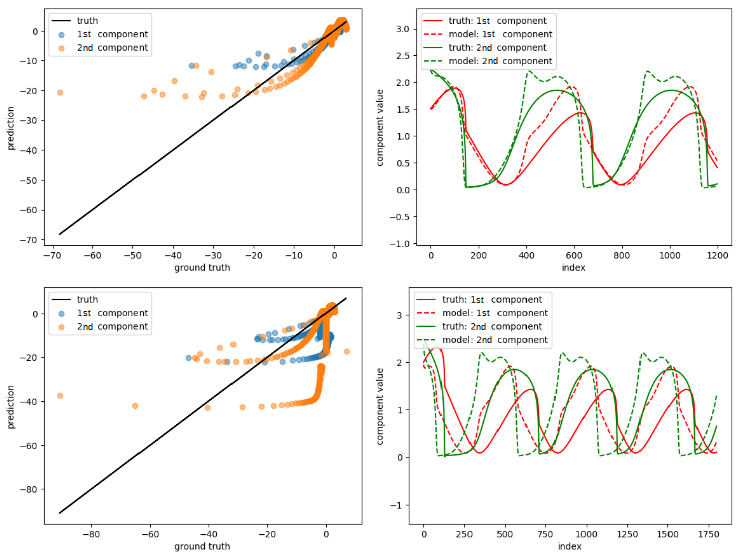}
	\caption{A priori and a posteriori result of SINDy on non-rational non-polynomial case. Top: training data. Bottom: testing data. Left: a priori. Right: a posteriori.}
	\label{fig:yg_validation_0_sindy}
\end{figure}

\begin{figure}[!htb]
	\centering
	\includegraphics[width=\textwidth]{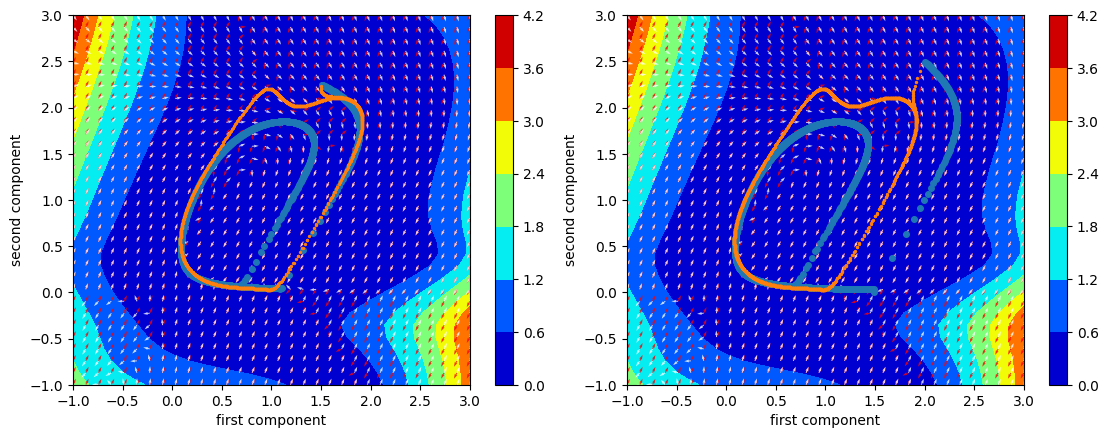}
	\caption{Stepwise error contour of SINDy on non-rational, non-polynomial case. Left: training data. Right: testing data. Blue dot: ground truth. Orange dot: prediction. White arrow: direction of target vector of ground truth. Red arrow: direction of target vector of prediction.}
	\label{fig:error_contour_YG_sindy}
\end{figure}

For SINDy, the polynomial order is set to three and threshold as $2\times 10^{-4}$. A priori and a posteriori validation for training and testing is shown in \cref{fig:yg_validation_0_sindy}. Correspondingly, the stepwise error contour displayed in \cref{fig:error_contour_YG_sindy} shows the misfit for the region of interests ranging from -1 to 3 for both two components. Because there is no sparsity in polynomial basis in this case, it is expected that SINDy cannot reconstruct the dynamics correctly and would perform worse than the basic model of FNN. \textcolor{black}{The implication is that for strongly non-polynomial systems,  neural networks are far more flexible compared to SINDy. }

\subsection{Nonlinear PDE system: flow behind a cylinder}\label{sec:cyd}

\begin{figure}[!htb]
	\centering
	\includegraphics[width=\textwidth]{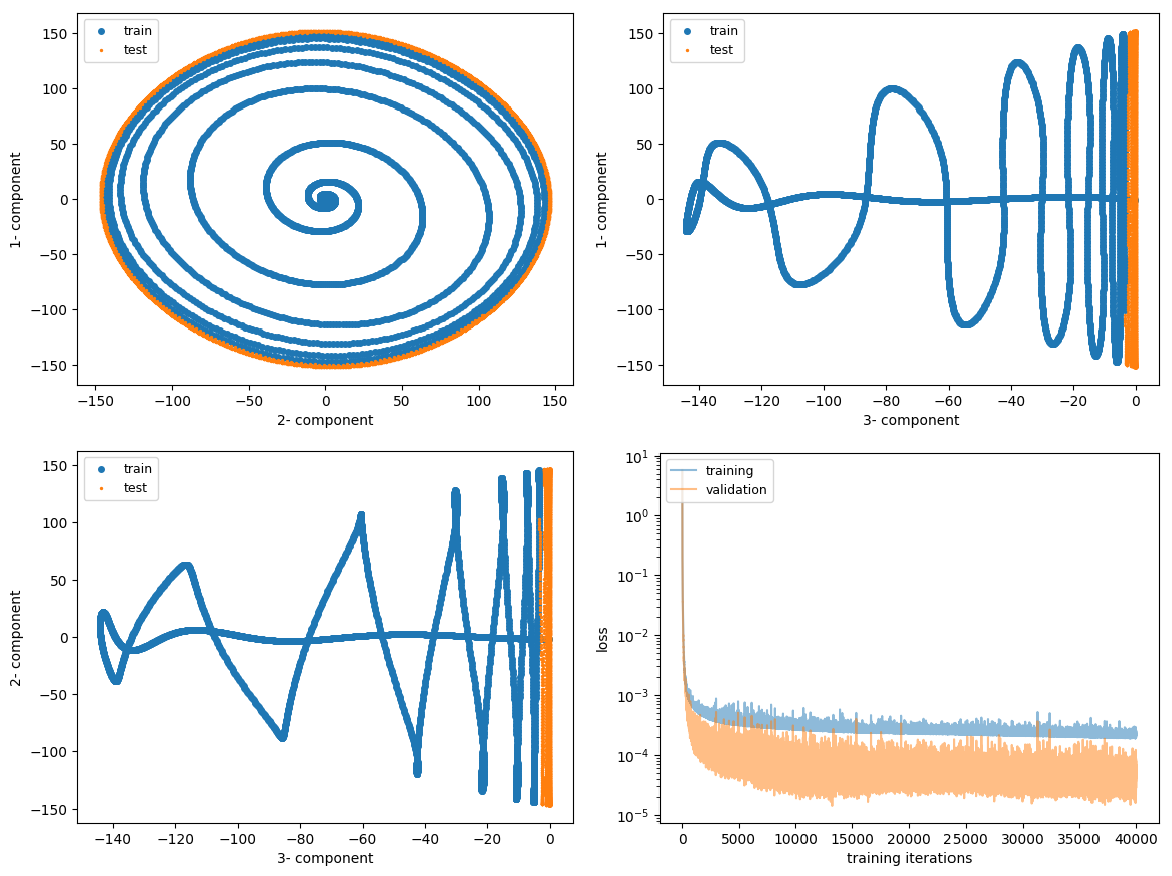}
	\caption{Flow in cylinder wake. Top left: data distribution of $x_1$ vs $x_2$. Top Right: data distribution of $x_1$ and $x_3$. Bottom left: data distribution of $x_2$ and $x_3$. Bottom right: learning curve.}
	\label{fig:cyd_train_target_dist_and_lc}
\end{figure}

In this section, we compare the basic model with SINDy in reconstructing the flow in a cylinder wake. 
\textcolor{black}{The
data is from Brunton et al.~\cite{Brunton2015} which comes from an immersed boundary method solution~\cite{taira2007immersed} of the 2D incompressible N-S equations with $Re =100$ based on the cylinder diameter. The computational domain consists of a non-uniform grid with near-wall refinement. The inlet condition is uniform flow and the outlet is a convective boundary condition to allow the vorticity to exit the domain freely.  Testing data is generated as a temporal extension of states that lie on a limit cycle at the boundary of training data, which indicates this is not an extrapolation task. To work with such a high-dimensional nonlinear PDE system, we use the coefficients of two POD modes~\cite{berkooz1993proper} and one `shift mode,’ which represents the shift of short-term averaged flow away from the POD space of the first two harmonic modes} to reduce the spatial dimension. \textcolor{black}{More details on POD and `shift-modes' are provided in Refs.~\cite{noack2003hierarchy,berkooz1993proper}.} Training and testing data is the same as in Brunton et al.~\cite{Brunton2015} where the first 2999 snapshots in time are used for training, and a later 2994 snapshots used for testing. A random 10\% of training snapshots is considered as validation set but also included in later evaluation. The distribution of training data and testing data is shown in \cref{fig:cyd_train_target_dist_and_lc}. 

\begin{figure}[!htb]
	\centering
	\includegraphics[width=\textwidth]{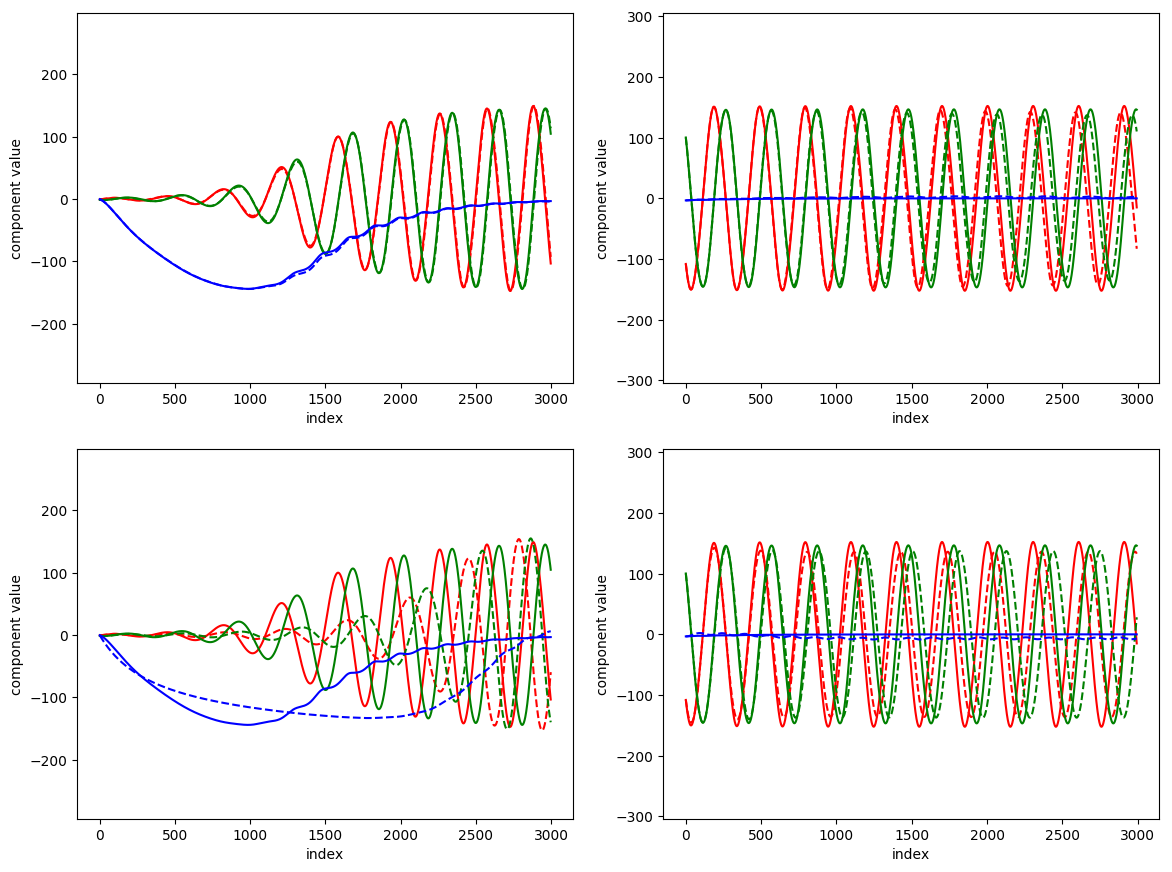}
	\caption{A posteriori comparison for flow in cylinder wake between basic model and SINDy. Red: first POD coefficient. Green: second POD coefficient. Blue: shift-mode coefficient. Top: basic model. Bottom: SINDy. Left: training data. Right: testing data. }
	\label{fig:compare_cylinder}
\end{figure}

Hyperparameters of the basic model are shown in \cref{table:hyperParameter_cyd_ann} with 40000 epochs. For SINDy, the hyperparameters are the same as in previous work \cite{Brunton2015}. As shown in \cref{fig:compare_cylinder}, for training data, SINDy reconstructs a smaller growth rate of oscillating behavior while the basic model accurately reconstructs both the shift mode and \textcolor{black}{two} POD modes. For testing data, SINDy contains an observable phase lag for the time period concerned, while the basic model achieves an almost perfect match. This implies that the model obtained from SINDy, although much easier to interpret than neural network, is not the best model for this dynamical system in terms of accuracy. \textcolor{black}{However, we note that from the data distribution in \cref{fig:cyd_train_target_dist_and_lc}, the basic model performs as expected, as the training data covers the attractor well.}

\begin{table}[!htb]
\caption{Hyperparameter configuration of basic model: flow in cylinder wake. }
\label{table:hyperParameter_cyd_ann}
\centering
\begin{tabular}{|c|c|c|c|c|} \hline
		layer structure & activation function & loss function & optimizer & learning rate  \\ \hline
		2-20-20-2 & elu & MSE & Adam & 0.001 \\  \hline
\end{tabular}
\end{table}

\subsection{Stabilizing the neural network with Jacobian regularization}\label{sec:jacobian}

Due to the non-convexity of the optimization problem that arises in the solution of the basic model in \cref{eq:basic_model}, employing a stochastic gradient-descent type method might lead to a solution corresponding to a local minimum, which is often undesirable and  difficult to avoid. Most works in the field of deep learning for feedforward neural networks focus on decreasing the impact of poor local minima to promote generalizability. However, in the context of modeling a dynamical system, as it is often assumed that the trajectory of interest is stable with respect to small disturbances \cite{Bakker2000}, the model should be able to approximately reconstruct the training trajectory in the presence of local errors that arise at each step. This would require regularizing instabilities that could arise in a posteriori prediction. To have meaningful comparisons, random number seeds are fixed for initialization of weights and training data shuffling. Nevertheless, we observe that in some cases, for example in the previous case of the cylinder wake, an inappropriate choice of neural network configuration of the basic model, e.g. number of hidden units and type of activation function, can potentially lead to instability in a posteriori evaluation. Such instabilities may materialize even while reconstructing the training trajectory,  while the corresponding a priori prediction is almost perfect. Previous work \cite{tsung1995phase} explicitly ensured stability by simply adding more \textcolor{black}{adjacent} trajectories. Here, we take a different approach by adding a \textcolor{black}{Jacobian} regularization term in the cost function in \cref{eq:regu_model}. 

\begin{figure}[!htb]
	\centering
	\includegraphics[width=\textwidth]{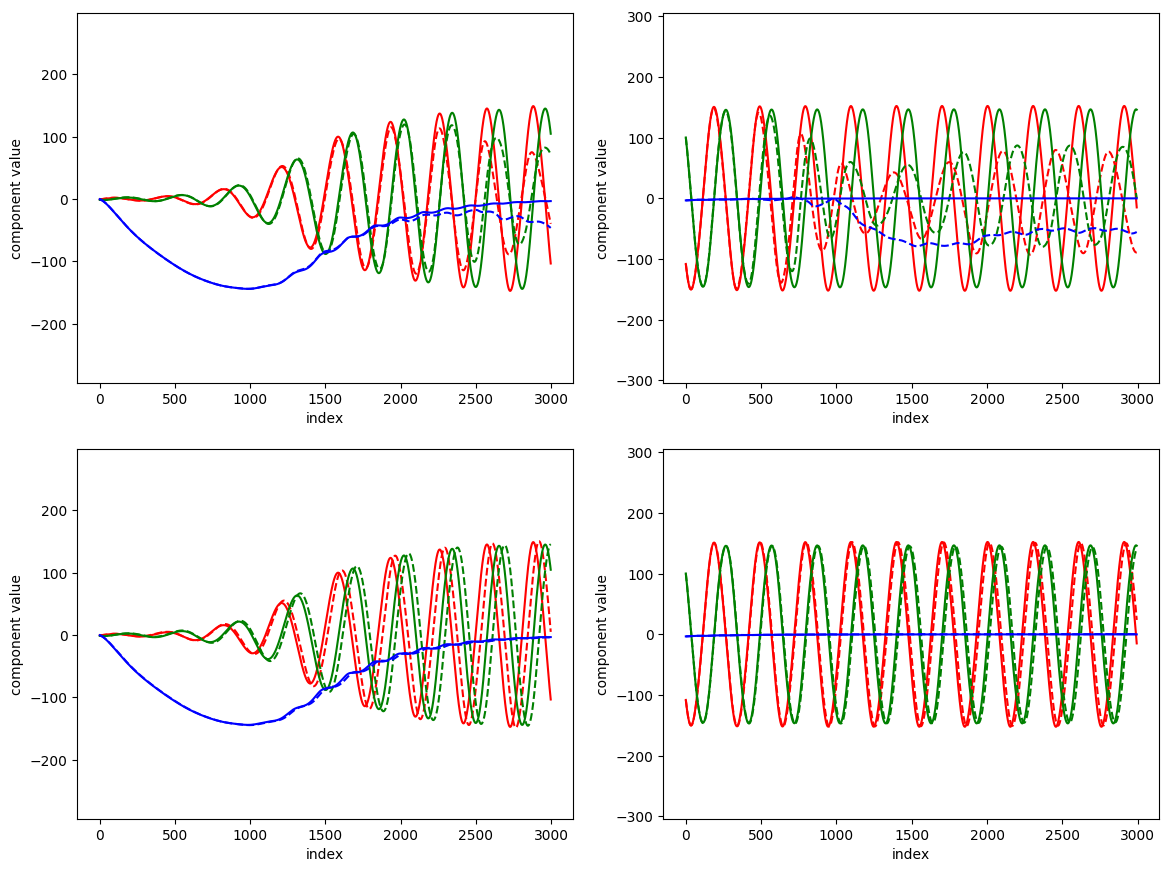}
	\caption{A posteriori comparison between basic and regularized models for cylinder wake. Red: first POD coefficient. Green: second POD coefficient. Blue: shift-mode coefficient. Top: basic model. Bottom: regularized model. Left: training data. Right: testing data. }
	\label{fig:compare_cylinder_regu}
\end{figure}

In our numerical experiments, with a certain fixed random seed, it is observed that, when the layer structure is 2-20-20-2 with $tanh$ as activation function instead of $elu$, the basic model becomes numerically unstable after 2000 steps for training data which is displayed in \cref{fig:compare_cylinder_regu}. Similar numerical instability is also observed in testing evaluations. However, for the same fixed random seed, the regularized model with $\lambda=5\times 10^{-5}$ shows numerically stable results with the same neural network configuration for both training and testing data. 

\begin{figure}[!htb]
	\centering
	\includegraphics[width=\textwidth]{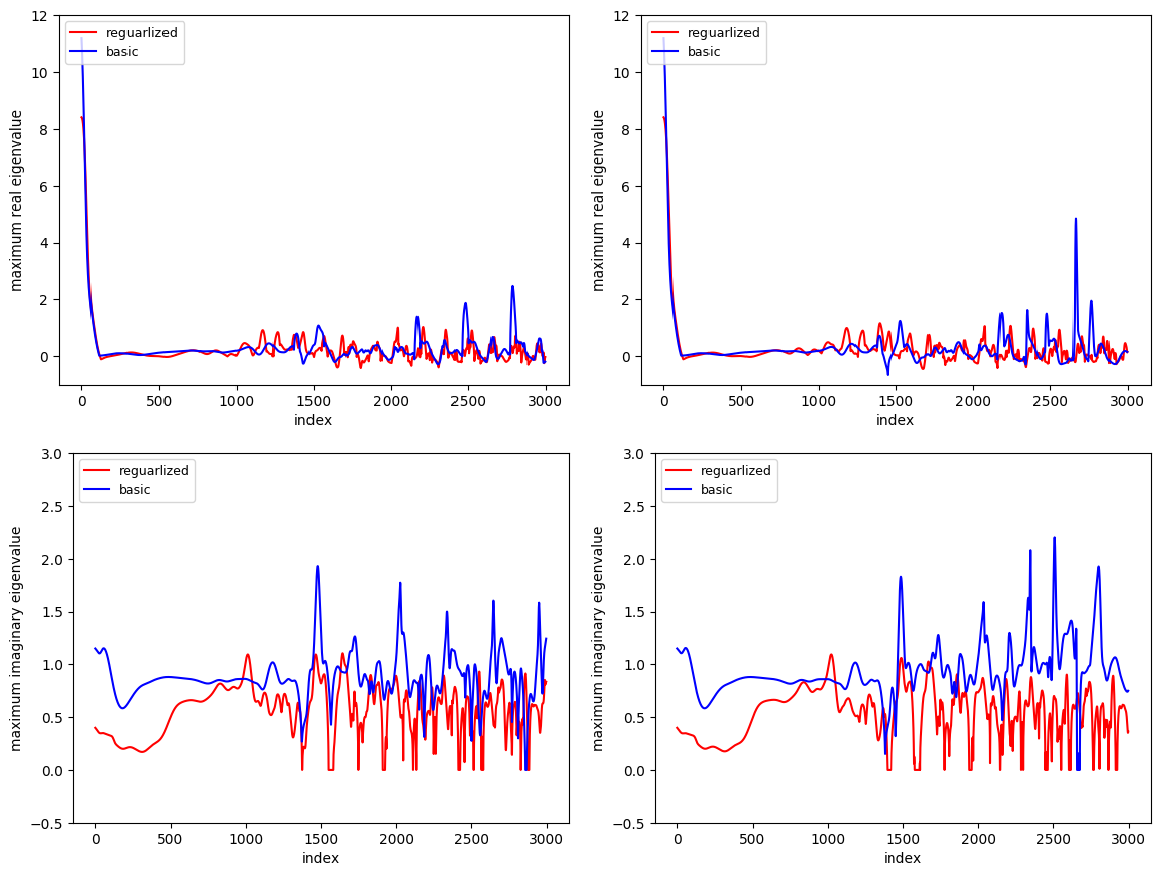}
	\caption{Flow in cylinder wake. Comparison of eigenvalue of Jacobian between regularized and basic model on training data. Top: maximal real eigenvalue. Bottom: maximal imaginary eigenvalue.  Left: a priori. Right: a posteriori. }
	\label{fig:CYD_regu_sv_training}
\end{figure}

\begin{figure}[!htb]
	\centering
	\includegraphics[width=\textwidth]{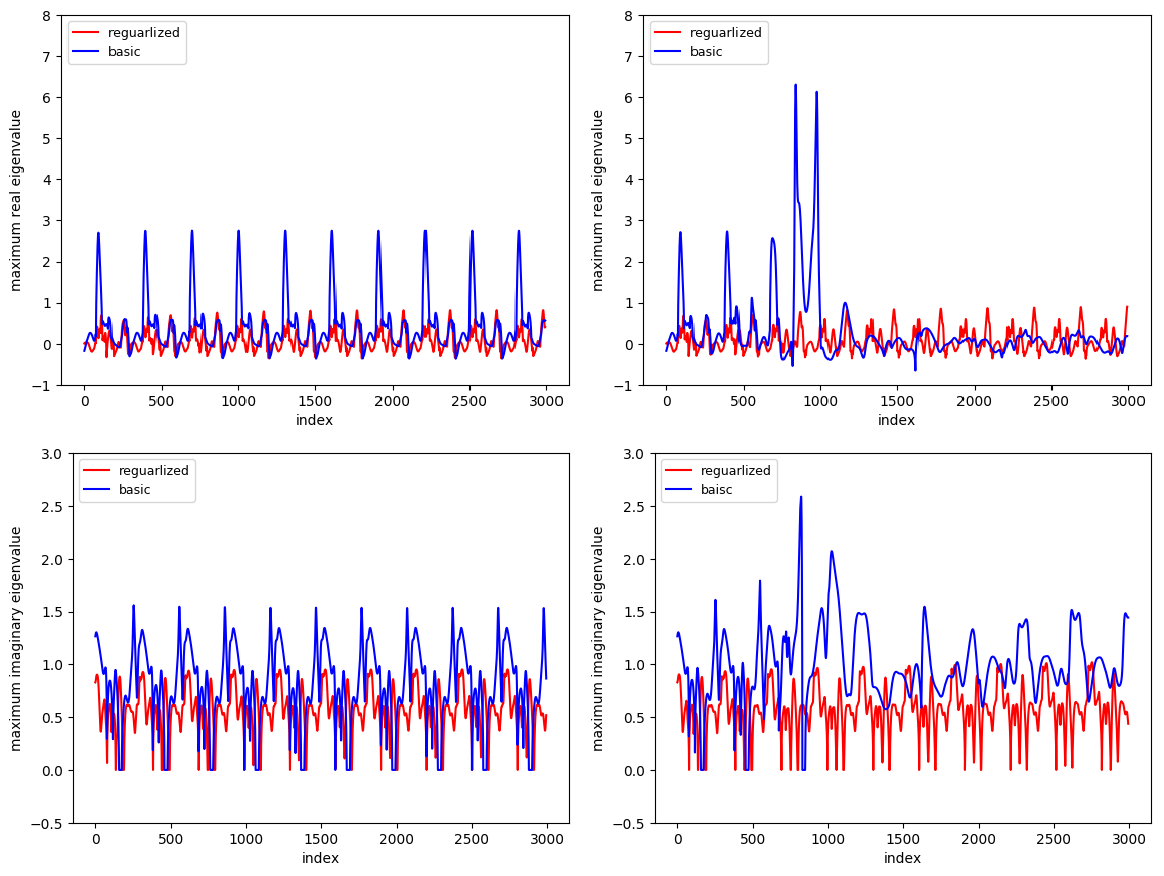}
	\caption{Flow in cylinder wake. Comparison of eigenvalue of Jacobian between regularized and basic model on testing data. Top: maximal real eigenvalue. Bottom: maximal imaginary eigenvalue.  Left: a priori. Right: a posteriori. }
	\label{fig:CYD_regu_sv_testing}
\end{figure}

\begin{figure}[!htb]
	\centering
	\includegraphics[width=\textwidth]{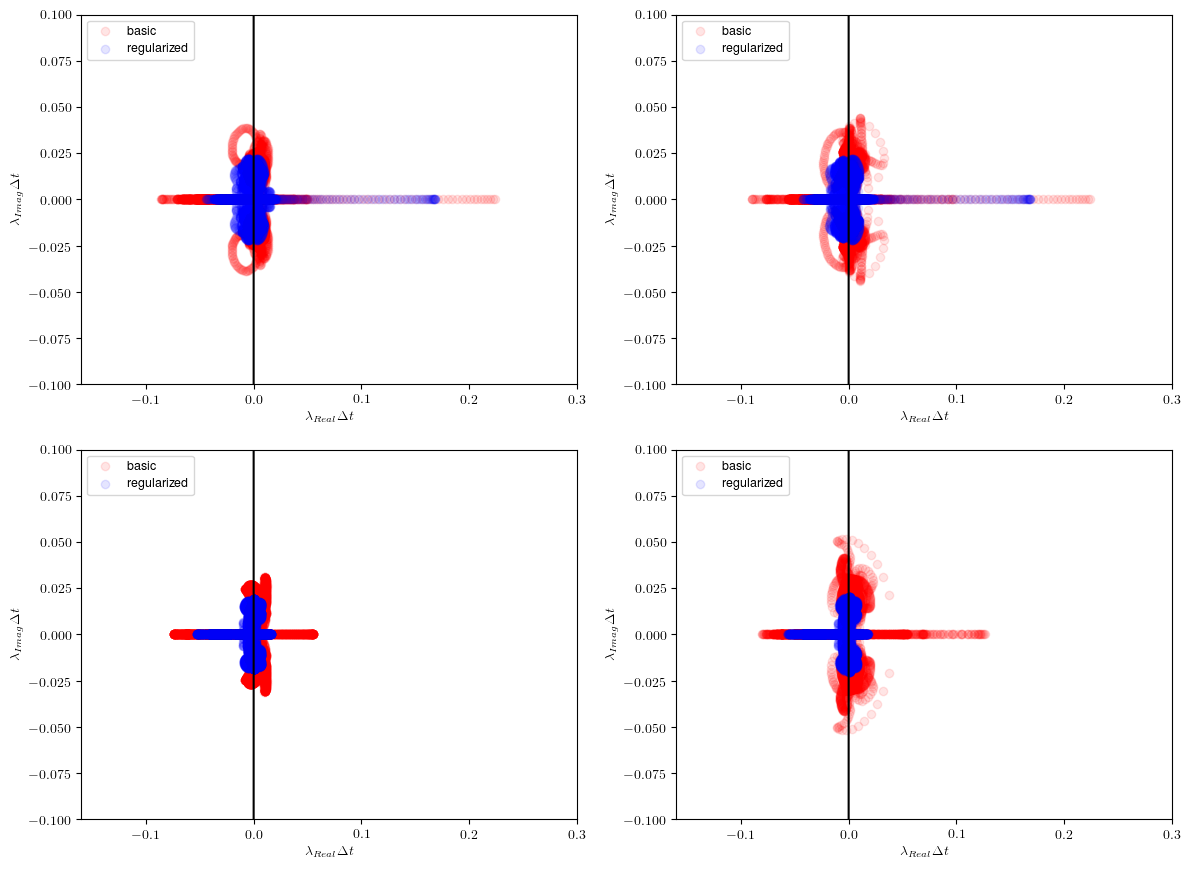}
	\caption{Flow in cylinder wake. Comparison of linear stability diagram between regularized and basic model. Top left: a priori on training data. Top right: a posteriori on training data. Bottom left: a priori on testing data. Bottom right: a posteriori on testing data.}
	\label{fig:CYD_linear_stab}
\end{figure}

The effectiveness of Jacobian regularization may be attributed to finding a balance between lowering the prediction error, i.e., MSE, and suppressing the sensitivity of the prediction of the future state to the current local error. As shown in \cref{fig:CYD_regu_sv_training} and \cref{fig:CYD_regu_sv_testing}, on average, the maximal eigenvalue of the Jacobian is smaller for the regularized model than for the basic model. 
Furthermore, the distribution of the eigenvalues of the Jacobian  is shown in \cref{fig:CYD_linear_stab} in the form of a linear stability diagram with explicit 5th order Runge-Kutta  time integration. It is  clear  that the model with Jacobian regularization has significantly smaller positive real eigenvalues. Note that, due to the Frobenius norm, negative real eigenvalues are also decreased in magnitude.


\subsection{Nonlinear PDE system: instability-driven buoyant mixing  flow}\label{sec:buoymix}

\textcolor{black}{The test problems thus far have served to assess the performance of the basic and Jacobian-regularized models on nonlinear dynamical systems that either evolve on or towards an attractor. Such systems, even if high-dimensional, are amenable for projection onto a lower dimensional subspace, using for instance, POD techniques.} In this section, we consider the Boussinesq buoyant mixing flow \cite{weinan98,liu03}, also known as the unsteady lock-exchange problem~\cite{san15} which exhibits strong shear and Kelvin-Helmholtz instability phenomena driven by the temperature gradient.  Compared to the cylinder flow that evolves on a low-dimensional attractor approaching a limit cycle, the Boussinesq flow is highly convective and instability driven. Consequently, such a system state cannot be represented by a compact set of POD modes \textcolor{black}{from the spatial-temporal field of nondimensionalized velocity and temperature.} Rather, the low-dimensional manifold itself evolves with time. 
Further, any noise in the initial data can produce unexpected deviations that makes such systems challenging to model, even using equation-driven reduced order models such as POD-Galerkin~\cite{san15}.

\begin{figure}[!htbp]
	\centering
	\includegraphics[width=\textwidth, scale=0.2]{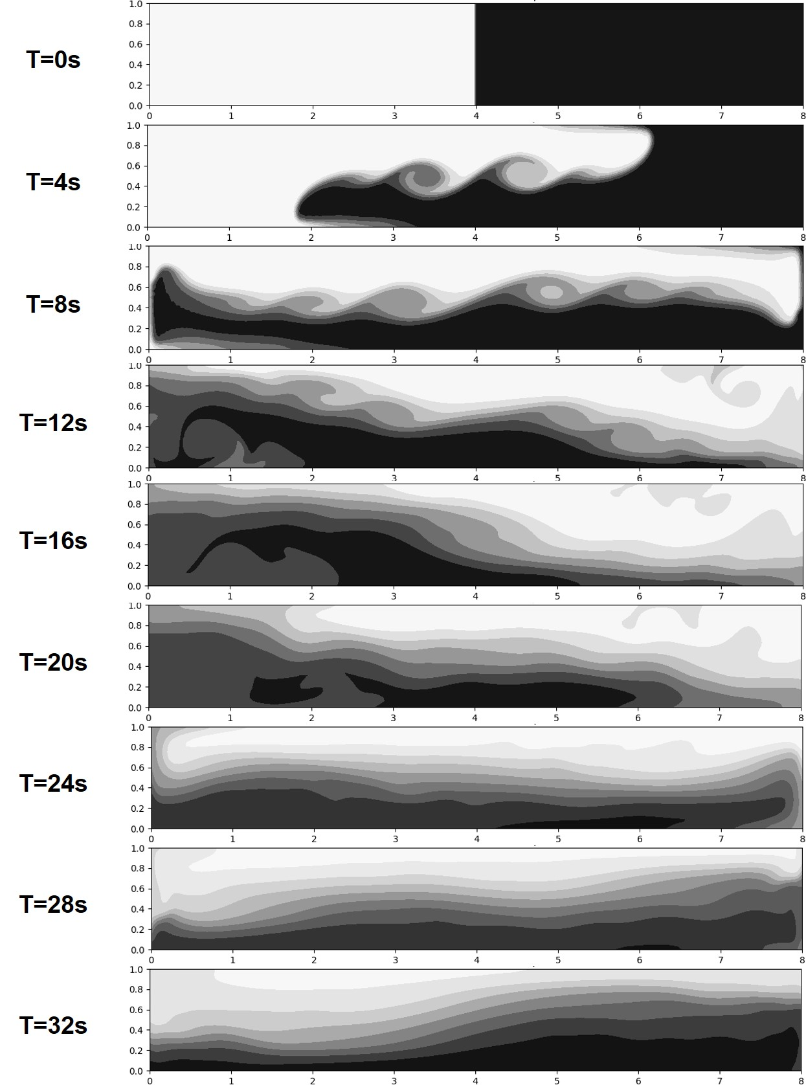}
	\caption{Time evolution of the temperature field for a two dimensional Buoyant mixing flow.}
	\label{fig:buoyantmixing_time_evolution}
\end{figure}

The data set is generated by solving the dimensionless form of the two-dimensional incompressible Boussinesq equations \cite{san15}, as shown in \cref{eq:bseq} on a rectangular domain that is $0<x<8$ and $0<y<1$.  
\begin{subequations} \label{eq:bseq}
\begin{equation}
\frac{\partial{u}}{\partial{x}}+\frac{\partial{u}}{\partial{y}}=0,
\end{equation}
\begin{equation}
\frac{\partial{u}}{\partial{t}}+u\frac{\partial{u}}{\partial{x}}+v\frac{\partial{u}}{\partial{y}}=-\frac{\partial{P}}{\partial{x}}+\frac{1}{Re}\nabla^2{u},
\end{equation}
\begin{equation}
\frac{\partial{v}}{\partial{t}}+u\frac{\partial{v}}{\partial{x}}+v\frac{\partial{v}}{\partial{y}}=-\frac{\partial{P}}{\partial{y}}+\frac{1}{Re}\nabla^2{v}+Ri\theta,
\end{equation}
\begin{equation}
\frac{\partial{\theta}}{\partial{t}}+u\frac{\partial{\theta}}{\partial{x}}+v\frac{\partial{\theta}}{\partial{y}}=\frac{1}{{Re}{Pr}}\nabla^2{\theta},
\end{equation}
\end{subequations}
where $u$, $v$, and $\theta$ are the horizontal, vertical velocity, and temperature components, respectively. The dimensionless parameters $Re$, $Ri$, and $Pr$ are the Reynolds number, Richardson number, and Prandtl number, respectively with values chosen as follows: $Re=1000$, $Ri=4.0$, and $Pr=1.0$. These equations are discretized on a $256 \times 33$ grid. Initially, fluids at two different temperatures are separated by a vertical line at $x=4$. The bounding walls are treated as adiabatic with the no-slip condition. A fourth-order compact finite difference scheme is used to compute the derivatives in \cref{eq:bseq}. The evolution of the thermal field over the simulation time interval of 32 seconds is shown in \cref{fig:buoyantmixing_time_evolution} and illustrates the highly transient nature of the dynamics. To reduce the dimensionality of the system, POD modes are extracted from the entire data set consisting of 1600 snapshots. The reduced feature set consisting of ten POD weights captures nearly 97\% of the total energy is used to train the model and predict the trajectory. 

For the setup of training and testing, future state prediction is pursued with the first 70\% states of the trajectory  treated as training data and the rest for testing. For such a system in 10 dimensions, it is observed that the problem of the a posteriori instability in the basic model becomes more pronounced and difficult to avoid. Challenges of numerical instability were observed even for reconstruction for a wide range of network configurations, and thus results from the basic model are not reported.

\begin{figure}[!htb]
	\centering
	\includegraphics[width=\textwidth, scale=1.0]{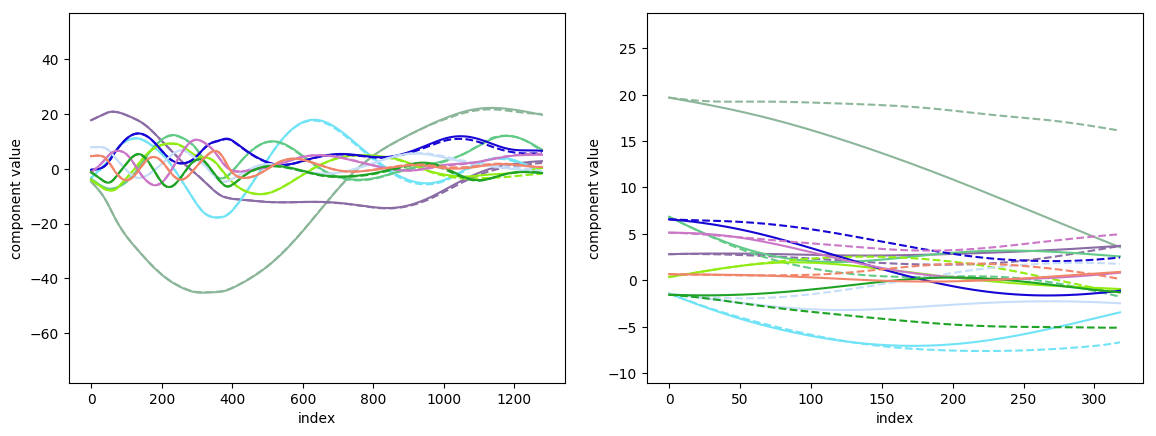}
	\caption{A posteriori comparison between prediction of Jacobian regularized model and ground truth for POD coefficient of buoyant mixing flow. Dashed: model. Solid: ground truth.
	First mode: \textcolor{color1th}{$\bullet$}. 
	Second mode: \textcolor{color2th}{$\bullet$}.
	Third mode: \textcolor{color3th}{$\bullet$}. 
	Fourth mode: \textcolor{color4th}{$\bullet$}.
	Fifth mode: \textcolor{color5th}{$\bullet$}. 
	Sixth mode: \textcolor{color6th}{$\bullet$}.
	Seventh mode: \textcolor{color7th}{$\bullet$}. 
	Eighth mode: \textcolor{color8th}{$\bullet$}.
	Ninth mode: \textcolor{color9th}{$\bullet$}. 
	Tenth mode: \textcolor{color10th}{$\bullet$}.}
	\label{fig:buoyantmixing_PODWeights_ML}
\end{figure}

The Jacobian regularized model is employed with hyperparameters shown in \cref{table:hyperParameter_byc_ann}, with  \cref{fig:buoyantmixing_PODWeights_ML} showing a posteriori evaluation on training data. The reconstruction is successful, but the performance deteriorates on testing data because the trajectory of the system does not exhibit a low dimensional attractor as in the cylinder case. Therefore the training data is not informative for predictions on the test set. For a black-box machine learning model, this phenomena can be expected to be more pronounced in high dimensional space due to data scarcity. Specifically, we discuss this problem in the following section. 


\begin{table}[!htb]
\caption{Hyperparameter configuration of Jacobian regularized model for buoyant mixing flow. }
\label{table:hyperParameter_byc_ann}
\centering
\begin{tabular}{|c|c|c|c|c|c|} \hline
		layer structure & activation function & loss function & optimizer & learning rate & $\lambda$ \\ \hline
		10-20-20-10 & penalized tanh \cite{xu2016revise} & MSE & Adam & 0.001 & 5e-4 \\  \hline
\end{tabular}
\end{table}

\subsection{Improving model predictability by data augmentation}\label{sec:info_rich}

In this section, we consider two scenarios of data augmentation: (i) augmenting the information in the data by spreading training locations randomly following a uniform distribution provided that one has access to $\bm{F}_{c}$ or $\bm{F}_d$ at any desired location; (ii) augmenting the data by assembling several trajectories generated from different initial conditions.

\subsubsection{Random uniform sampling in phase space}


Recall that, in the two-dimensional problems in \cref{sec:result_vdp} and \cref{sec:yg}, the stepwise error contour shows that local error increases on testing scenarios located far away from the training data which was highly concentrated in a compact region of phase space. Without any knowledge of system behavior, it is sensible to start with training data from a random uniform distribution in a compact region of phase space corresponding to interesting dynamics. 
To conduct a thorough stepwise error contour evaluation of the training target in phase space, the VDP system is chosen to illustrate this idea.


Determining the most informative data samples would potentially involve  specific knowledge of the underlying system and the models used, and is beyond the scope of current work. Here we simply consider uniform random sampling in the phase space in a finite domain: $[-3,3]$ for the first component and $[-5,5]$ for the second component. We obtained a new set of 399 training data points using random uniform sampling in phase space while retaining the same testing data as in \cref{sec:result_vdp}.

\begin{figure}[!htb]
	
	\centering
	\includegraphics[width=\textwidth]{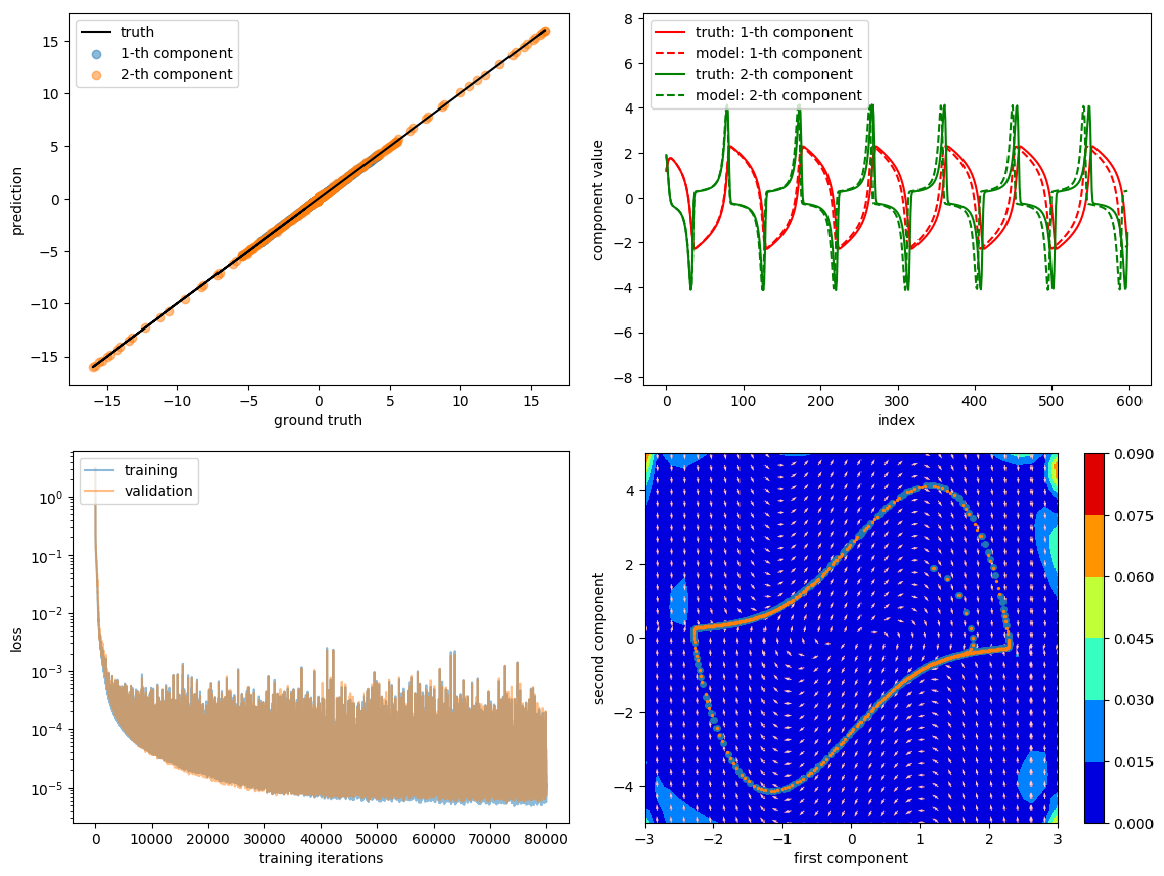}

	\caption{A priori and a posteriori result of the basic model on the VDP system, training with random sampled data distribution. Top left: a priori evaluation. Top right: a posteriori evaluation. Bottom left: learning curve. Bottom right: stepwise error contour. Blue dot: ground truth. Orange dot: prediction. White arrow: direction of target vector of ground truth. Red arrow: direction of target vector of prediction.}
	\label{fig:model_performance_data_rich}
\end{figure}

The performance of the basic model with the same hyperparameter setting as in \cref{sec:result_vdp} on randomly distributed training data is shown in \cref{fig:model_performance_data_rich}. While the number of data points has not been changed, the contour error of the resulting model decreased significantly compared to  training with the same number of data points in a single trajectory, which indicates an improved generalizability with the same amount of training data.

\subsubsection{Training with multiple trajectories with random initialization}\label{sec:vbe}


Training data can also be augmented by multiple trajectories with different initial conditions. Here we take the one-dimensional viscous Burgers equation shown in \cref{eq:vbe} as example.



The initial conditions are generated following a specific energy spectrum \cite{li2016priori,Parish2016} shown in \cref{eq:spectrum}. 
\begin{equation}{\label{eq:vbe}}
\frac{\partial  u}{\partial t} + u \frac{\partial u }{\partial x} = \nu \frac{\partial^2 u}{\partial x^2},
\end{equation}
where $x \in [0, 2\pi]$ is a periodic domain discretized using 2048 uniformly distributed grid points, and $t\in[0, 20]$, $\nu=0.01$.
\begin{equation}{\label{eq:spectrum}}
    u(x,0) = \sum^{k_c}_{k=1} \dfrac{1}{\pi} \sqrt{2AE(k)} \sin(k x + \beta_k ),
\end{equation} where for each $k$, $\beta_k$ is a random number drawn from a uniform distribution on $[-\pi, \pi]$, $E(k) = 5^{-5/3}$ if $1 \le k \ge 5$, $A=25$, and $E(k) = k^{-5/3}$ if $k > 5$. Multiple trajectories are generated using different seeds for random numbers to obtain the trajectories of the full-order system. To fully resolve the system as a DNS, \cref{eq:vbe} is solved using a standard pseudo-spectral method with SSP-RK3 \cite{gottlieb2001strong} for time stepping. Here we choose $k_c = 2$. Discrete cosine transformation (DCT) is used to reduce the dimension of the full system to first 4 cosine modes in the system where around 97\% of kinetic energy is preserved. For simplicity, we  seek a closed Markovian reduced-order-system, whereas the underlying dynamics is clearly non-Markovian~\cite{pan2018data,Parish2016}.

\begin{figure}[!htb]
	\centering
	\includegraphics[width=\textwidth]{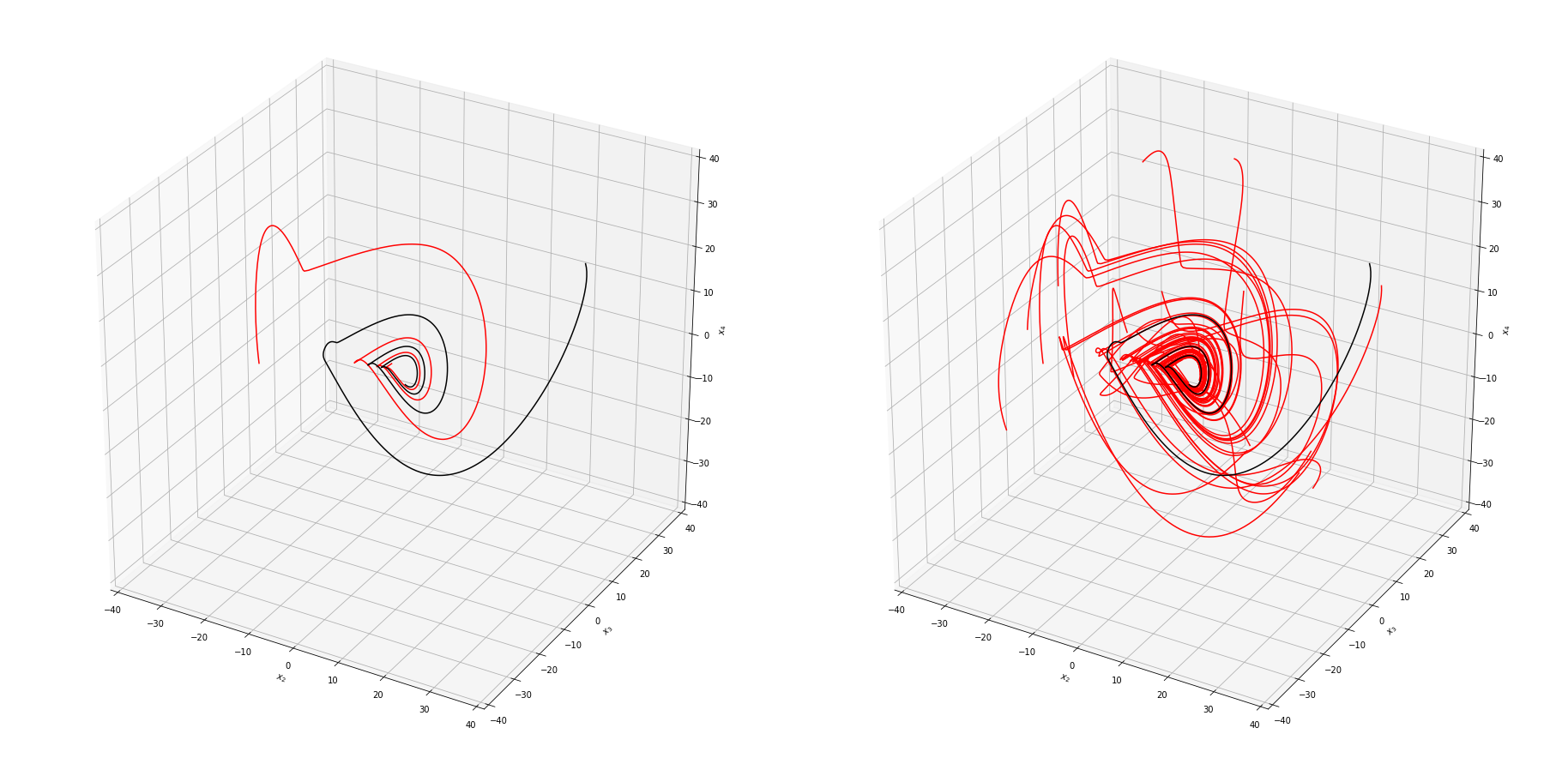}
	\caption{Feature distribution of training/testing data for component $x_2, x_3, x_4$. Left: with one training trajectory. Right: with 18 training trajectory. Black line: testing data. Red line: training data.}
	\label{fig:vbe_train_feature_dist_and_lc_3d}
\end{figure}

Since the first component of the DCT is constant, the remaining components of  feature space are shown in \cref{fig:vbe_train_feature_dist_and_lc_3d}. The training data is far away from the testing data initially, whereas the data converges at a later stage. This is because of the presence of a spiral fixed point attractor resulting from the viscous dissipative nature of the system. Therefore, if the model is only trained  from a single trajectory, it will be very difficult for the model to generalize well in the phase space especially where the state of the system is not near an attractor. 

Many dynamical systems in nature exhibit attractors in the asymptotic sense. From the viewpoint of data-driven modeling of such dynamics, data scarcity is encountered at the start of trajectory where the number of trajectories required  to provide enough information to cover the region of interest grows exponentially. Much research on applying neural network-based models for  dynamical systems \cite{tsung1995phase}\cite{Smaoui1997a}\cite{Wang2017} demonstrate problems starting on limit cycles or chaotic attractors in a low-dimensional feature space, where the issue of initial data scarcity is not significant, or can be easily alleviated by a small increase in available data. However, for the purpose of modeling phenomena such as turbulent fluid flow, which can be  high dimensional even after dimension reduction, the model would likely  fail for long-time prediction due to data scarcity. Such a situation may be realized in regions of phase space where the state has not arrived at the low manifold attractor. Therefore, the training data might not be representative of testing data which violates the fundamental assumption of a well-posed machine learning problem ~\cite{friedman2001elements}. Moreover, data scarcity will shrink the region of generalizability of the model as the dimension of the system increases. 

A key benefit of using a neural network model is its linear growth in complexity with dimension of the system, in contrast  to traditional polynomial regression methods \cite{goodfellow2016deep}. However, initial data scarcity would limit the  generalizability of a ANN in modeling a high dimensional dynamical system that does not exhibit a low dimensional attractor. We believe this phenomenon of data scarcity observed from this simple nonlinear PDE example also applies to other nonlinear dynamical systems.

To alleviate the initial data scarcity issue,  a  solution is to augment the training data with more trajectories with different random number seeds in generating the initial condition, while keeping the energy spectrum the same across all cases. In this case we choose 18 such trajectories. Each trajectory contains states of 1000 snapshots equally spaced in time. For testing data, we simply consider one DNS result with an initial condition different from all training trajectories. The corresponding training and testing trajectories are visualized in phase space as shown in \cref{fig:vbe_train_feature_dist_and_lc_3d}.

\begin{figure}[!htb]
	\centering
	\includegraphics[width=\textwidth]{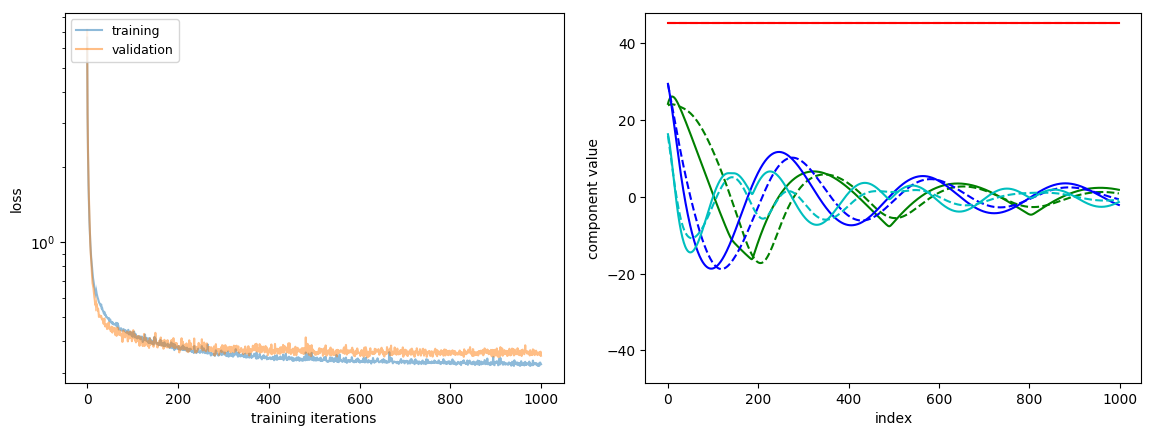}
	\caption{Result of basic model for one dimensional VBE turbulence case. Left: learning curve. Right: a posteriori result with augmented training trajectories. Red: $x_1$. Green: $x_2$. Blue: $x_3$. Cyan: $x_4$. Dashed: model prediction. Solid: ground truth.}
	\label{fig:vbe_train_test_posteriori_ann}
\end{figure}

The basic model is trained with hyperparameters in \cref{table:hyperParameter_vbe_ann} and 1000 epochs. The resulting learning curve and a posteriori evaluation are shown in \cref{fig:vbe_train_test_posteriori_ann}. Relatively large discrepancy is observed near the initial condition as the initial data scarcity is not completely eliminated due to limited number of additional trajectories. Increasing the number of additional trajectories, may be unaffordable for very high dimensional systems. Moreover, the result also shows that the error decreases once the trajectory falls on the fixed point attractor. Thus, if the model starts in the low dimensional attractor where the information is well-preserved in the training data, better performance might be expected. This hypothesis is consistent with previous work~\cite{Wang2017}, where successful  prediction of future states starts at the time when the states converge to a low dimensional attractor.

\begin{table}[!htb]
\caption{Hyperparameter configuration of basic model: VBE system with four modes}
\label{table:hyperParameter_vbe_ann}
\centering
\begin{tabular}{|c|c|c|c|c|} \hline
		layer structure & activation function & loss function & optimizer & learning rate  \\ \hline
		4-30-30-30-4 & elu & MSE & Adam & 0.0005 \\  \hline
\end{tabular}
\end{table}

%



\section{Conclusions}
\label{sec:conclusions}

This work investigated the modeling of dynamical systems using feedforward neural networks (FNN), with a focus on long time prediction. It was shown that neural networks have  advantages over sparse polynomial regression in terms of adaptability, but with a trade-off in training cost and \textcolor{black}{difficulty in extrapolation, which is a natural barrier for almost all supervised learning.} From the perspective of global error analysis, and the observation of the strong correlation between the local error and maximal singular value of the Jacobian, we propose the suppression of the Frobenius norm of the Jacobian as regularization. This showed promise in improving the robustness of the basic FNN model given limited data, or when the model has a non-ideal architecture, or when the model is unstable. The effectiveness of Jacobian regularization is  attributed to finding a balance between lowering the prediction error, and suppressing the sensitivity of the prediction of the future state to the current local error.  In terms of modeling dynamical systems that do not involve low-dimensional attractors,  limitations of FNNs, and perhaps all local ML methods, was demonstrated in a buoyant mixing flow. Challenges were noted in the example of the reduced-order viscous burgers system, where significant initial data scarcity is present. Augmenting the data either by altering the distribution of training data in phase space or by simply adding multiple trajectories from different initial conditions resulted in improvement of the performance of FNN model to some extent. However, these remedies require a significant amount of additional sampling in phase space, especially for high dimensional systems \textcolor{black}{for the period of time without the apparent low-dimensional attractor} which suffers \textcolor{black}{data sparsity} from the curse of dimensionality.

\begin{acknowledgements}
The authors acknowledge Professor Balaji Jayaraman for helpful discussion, comments and providing the data for the two dimensional instability-driven buoyant mixing flow. This work was supported by AFOSR and AFRL under grants FA9550-16-1-0309 \& FA9550-17-1-0195.
\end{acknowledgements}

\bibliographystyle{plain}      
\bibliography{references}   


\end{document}